\documentclass[english,a4paper,12pt]{article}

\usepackage{amsxtra, amsfonts, amstext, amsmath, mathtools}%, 
\usepackage{amsthm}
\usepackage{dsfont}
\usepackage{booktabs}
\usepackage{nicefrac}
\usepackage{xspace}
\usepackage{url}
\usepackage{graphics,color}
\usepackage[algo2e,ruled,vlined,linesnumbered]{algorithm2e}
\usepackage{wrapfig}

\usepackage{color}

\newtheorem{theorem}{Theorem}

\DeclareMathOperator{\poly}{poly}
\newcommand{\oea}{\mbox{$(1 + 1)$~EA}\xspace}

\newcommand{\oplea}{\mbox{$(1+\lambda)$~EA}\xspace}

\newcommand{\mclea}{\mbox{${(\mu,\lambda)}$~EA}\xspace}
\newcommand{\oclea}{\mbox{$(1,\lambda)$~EA}\xspace}
\newcommand{\opllga}{\mbox{$(1+(\lambda,\lambda))$~GA}\xspace}
\newcommand{\ollga}{\mbox{$(1+(\lambda,\lambda))$~GA}\xspace}

\newcommand{\LB}{\textsc{LB}\xspace}
\newcommand{\onemax}{\textsc{OneMax}\xspace}
\newcommand{\LO}{\textsc{Leading\-Ones}\xspace}
\newcommand{\leadingones}{\LO}

\DeclareMathOperator{\Sample}{Sample}
\newcommand{\assign}{\leftarrow}

\newcommand{\R}{\ensuremath{\mathbb{R}}}

\newcommand{\N}{\ensuremath{\mathbb{N}}} % ohne Null!!!

\newcommand{\eps}{\varepsilon}

\newcommand{\gsemo}{\textsc{GSEMO}\xspace}
\newcommand{\gsemoc}{\textsc{GSEMO-C}\xspace}
\newcommand{\pomc}{\textsc{POMC}\xspace}

\let\originalleft\left
\let\originalright\right
\renewcommand{\left}{\mathopen{}\mathclose\bgroup\originalleft}
\renewcommand{\right}{\aftergroup\egroup\originalright}

\usepackage{hyperref}

\begin{document}

\title{A Survey on Recent Progress 
in the Theory of Evolutionary Algorithms for Discrete
Optimization}

\author{Benjamin Doerr\thanks{Laboratoire d'Informatique (LIX), CNRS, \'Ecole Polytechnique, Institut Polytechnique de Paris, Palaiseau, France} \and Frank Neumann\thanks{Optimisation and Logistics, School of Computer Science, The University of Adelaide, Adelaide, South Australia, Australia}}

\maketitle

\begin{abstract}
The theory of evolutionary computation for discrete search spaces has made significant progress in the last ten years. This survey summarizes some of the most important recent results in this research area. It discusses fine-grained models of runtime analysis of evolutionary algorithms, highlights recent theoretical insights on parameter tuning and parameter control, and summarizes the latest advances for stochastic and dynamic problems. We regard how evolutionary algorithms optimize submodular functions and we give an overview over the large body of recent results on estimation of distribution algorithms. Finally, we present the state of the art of drift analysis, one of the most powerful analysis technique developed in this field.
\end{abstract}

%%
%% This command processes the author and affiliation and title
%% information and builds the first part of the formatted document.

{\sloppy
\section{Introduction}

Evolutionary computing techniques have been applied in a large variety of different settings ranging from classical optimization problems in the context of supply chain management and renewable energy~\cite{DBLP:series/sci/BonyadiM16,DBLP:conf/gecco/TranWDA0N13,DBLP:journals/isci/NeshatAW20} over to the creation of music and art~\cite{DBLP:series/isrl/Dostal13b,DBLP:conf/ncs/Lewis08,DBLP:journals/corr/abs-2003-01517}.
The easy applicability of evolutionary algorithms makes them attractive also to users from outside of computer science disciplines and is one of the major reason for their success in a wide range of engineering applications such as the design of water networks~\cite{DBLP:journals/envsoft/BiDM15} or processing and planning in mining~\cite{DBLP:conf/gecco/MyburghD10,DBLP:conf/cec/OsadaWBM13,DBLP:journals/corr/abs-2102-05235}.

The theoretical understanding and analysis of evolutionary algorithms is key to further improve the applicability and performance of evolutionary computing methods in a wide range of settings. The area of runtime analysis has played a predominant role during the last 25 years in the area of theory of evolutionary computation when considering discrete optimization problems. 
This area provides rigorous insights based on mathematical analyses of the search behaviour of evolutionary algorithms. Such theorem-proof style results do not depend on the design of experimental settings which often comes with problems regarding interpretability. While the mathematical results obtained here are often for more restricted settings than in experimental studies, their proofs not only guarantee the result to be correct, but also show why a certain algorithm shows a certain optimization behavior. This way, theoretical research can provide a deeper way of understanding how evolutionary algorithms work.

The goal of this survey is to point out important research directions and their main results of the last 5-10 years. Due to the size of this research area, this survey cannot give a complete description of the progress in this field. We therefore focus on six areas that from our (subjective) point of view are currently the most active and promising ones.

While the area of rigorous runtime analysis provides theoretical insights based on mathematical proofs, it should be noted that there are other approaches contributing to the theoretical understanding of evolutionary algorithms. In particular, important insights into the working behaviour of evolutionary algorithms can be gained through sound experimental studies which investigate important modules and parameters of evolutionary algorithms~\cite{DBLP:series/ncs/Bartz-BeielsteinP14,DBLP:journals/biodatamining/SipperFAM18}. Furthermore, the area of fitness landscape analysis~\cite{DBLP:series/sci/PitzerA12,DBLP:conf/gecco/KerschkeP19} has contributed significantly to the theoretical understanding of evolutionary computing methods in relation to characteristics of the landscape of optimization problems such as the classical Traveling Salesperson problem~\cite{DBLP:journals/tcs/WhitleySOC14,DBLP:journals/heuristics/OchoaV18}. Feature-based analysis is another technique which tries to capture features of an optimization problem that make it hard or easy to be optimised by a given algorithm~\cite{DBLP:journals/amai/Smith-MilesH11,DBLP:journals/amai/MersmannBT0BN13,DBLP:journals/firai/Nallaperuma0N15}. All these areas make important contributions to the theoretical understanding of evolutionary computing and set the basis for the novel design of high performing evolutionary algorithms. We concentrate in this survey on results obtained in the area of runtime analysis and recommend the reader to consult the original articles for the previously named areas.

We now give an overview on the different areas covered in this survey.
We start with approaches that provide a \emph{fine-grained runtime analysis} of evolutionary computing in Section~\ref{sec:param}. This area investigates the runtime with respect to the given input size and additional parameters such as a fixed target value or results achievable within a fixed budget. 
Another important area that provides a fine-grained view is the area of parameterized runtime analysis which takes into account important structural parameters of a given problem instance. These approaches allow one to give a more fine-grained view on the runtime behavior and reveals how structural parameters or computational budgets influence the results. The parameterized analyses carried out in this area are focused on classical combinatorial optimization problems such as minimum vertex cover and the Euclidean Traveling Salesperson problems and we will summarize the main results for them.

Setting the \emph{parameters of evolutionary algorithms} is a key challenge in the profitable use of these heuristics. In Section~\ref{sec:parameters}, we discuss how recent theoretical works suggest to set the parameters. We also discuss different ways to let the algorithm optimize its parameters itself, which currently appears as a very powerful, easy-to-use approach.

\emph{Dynamic and stochastic problems} play a key role in many real-world applications~\cite{DBLP:series/sci/RitzingerP13,DBLP:journals/corr/abs-2102-05235,DBLP:conf/gecco/XieN021} and evolutionary algorithms have been shown to be very successful in dynamic and stochastic environments~\cite{DCOPS,DBLP:journals/swevo/RakshitKD17}. The theoretical investigations in terms of runtime analysis for such problems have been started by Droste~\cite{Droste02,DBLP:conf/gecco/Droste03,DBLP:conf/gecco/Droste04} in the early 2000s and a wide range of results have been obtained during the last 10 years. We will summarize such results in Section~\ref{sec:dynamic}.

Many important problems can be formulated in terms of a \emph{submodular functions} with a given set of constraints. The analysis and the design of evolutionary algorithms for submodular optimization problems has gained a lot of attention during the last 5 years. Various types of constraints as well as dynamic and stochastic settings have been investigated and provably efficient evolutionary algorithms outperforming previous state-of-the-art greedy approaches have been designed. The most important results and the different areas investigated are presented in Section~\ref{sec:submodular}.

\emph{Estimation-of-distribution algorithms} (EDAs) are evolutionary algorithms which do not evolve a population of good solution candidates, but a probability distribution on the search space that allows one to sample good solutions. Due to the complicated nature of the underlying mathematical objects (a random process taking probability distributions as states), for a long time the theoretical understanding of these algorithms was very limited. The last few years, however, have seen great progress in this topic, both showing new advantages of EDAs such as robustness to noise and giving advice in how to set their parameters. We review some of these results in Section~\ref{sec:eda}.

We end this survey with a discussion of an important analysis tools developed in the EA community. The area of \emph{drift analysis} has provided a wide range of analytical methods that allow to translate an understanding of the typical progress of an EA into information about its runtime. We will summarize the most important drift theorems and their applications together with the challenges involved when using drift analysis in Section~\ref{sec:drift}.

%\section{Methods for the Analysis}
%(Benjamin)
%Probabilistic Tools for the Analysis of Randomized Optimization Heuristics

%\section{Computational Complexity Settings}

%Complexity Theory for Discrete Black-Box Optimization Heuristics

\section{Fine-grained Runtime Analysis of Randomized Search Heuristics}
\label{sec:param}

Traditional runtime analysis investigates the runtime of an evolutionary computing technique on a given problem with respect to the size of problem instance. Usually, a worst-case view is taken and upper bounds on the runtime of all problem instances of a given size are proven. 

Many important optimization problems can be stated as binary problems.
A binary problem is defined on the search space $\{0,1\}^n$ and a problem instance is given as a pseudo-Boolean functions $f \colon \{0,1\}^n \rightarrow \mathds{R}$. We call $n$ the size of the instance. A common goal is to analyse the expected number of solutions that an evolutionary algorithm generates until an optimal solution with respect to $f$ has been produced for the first time. In general, the runtime is measured in terms of the number of solutions that are produced and evaluated before the considered algorithm has obtained a given goal for the first time.
The \emph{expected optimization time} refers to the expected number of constructed solutions until an optimal solution has been produced for the first time.

Studies focus on simplified algorithms such as the (1+1)~EA which is a special case of the ($\mu+1$)~EA shown in  Algorithm~\ref{alg:paramcomplex:muone} for maximizing a pseudo-Boolean function $f$. Randomized local search (RLS) is an even simpler variant and differs from the (1+1)~EA by flipping in each iteration exactly $1$ bit chosen uniformly at random. For problems where $1$-bit flips leads to obvious local optima a combination of $1$-bit and $2$-bit flips is considered~\cite{NeumannW07}.

Early analyses reveal that the expected time until RLS and the classical (1+1)~EA has obtained an optimal solution for the maximization of the benchmark function 
$$OneMax(x) = \sum_{i=1}^n x_i$$
is $\Theta(n \log n)$. Similarly, the expected runtime of RLS and the (1+1)~EA for the maximization of the benchmark function
$$
LeadingOnes(x) = \sum_{i=1}^n \prod_{j=1}^i x_j
$$
is $\Theta(n^2)$. Studies of this type have been frequently carried out in the area of runtime analysis over the last $25$ year.
Furthermore, the research area discusses the time to obtain an optimal solution or the time to achieve a good approximation in the case of NP-hard problems. One drawback of the described approach is that it is not able to distill other important characteristics that make a problem instance hard or easy to solve. 

 During the last years, different approaches have been taken to give a more precise view on the runtime behavior of evolutionary algorithms. This includes approaches for evaluating the runtime until a fixed target value (instead of an optimal value) is achieved. Furthermore, the parameterized analysis allows to obtain results in terms of additional structural parameters.
We start by summarizing some important results in the area of fixed budget and fixed target analysis and discuss parameterized results afterwards.

\begin{algorithm2e}[t]
  \SetKwFor{For}{repeat}{}{}
  Choose a population $P$ consisting of $\mu$ solutions chosen from $\{0,1\}^n$ uniformly at random\;
  \For{forever}{%
    Choose $x\in P$ randomly\;
    Create $x'$ by flipping each bit of $x$ independently with probability $1/n$\;
 %   Determine $f(x')$\;
    Add $\{x'\}$ to $P$.\;
    Remove an element $y$ from $P$ for which $y = \arg \min_{x \in P} f(x)$ holds\; 
  }
\caption{($\mu+1$)-EA for maximization problems}
\label{alg:paramcomplex:muone}
\end{algorithm2e}

\subsection{Fixed Budget and Fixed Target Analysis}
Fixed budget analysis has been introduced in \cite{DBLP:conf/gecco/JansenZ12} and considers the quality that an evolutionary algorithm achieves within a given time budget $b$. Classical runtime analysis often already provides the insights for such fixed budget analysis as it analyzes the progress that an evolutionary algorithm makes at the different stages of the optimization process. Taking a more fine-grained view on such classical runtime results, fixed budget results can often be obtained. In \cite{DBLP:conf/gecco/JansenZ12}, results for RLS and the (1+1)~EA on the benchmark problems OneMax and LeadingOnes are obtained which are based on more fine-grained insights of previous studies on the expected optimization time. An extension of these fixed budget results has been provided in \cite{DBLP:journals/tcs/JansenZ14}.
Furthermore, a general method for obtaining fixed budget results based on classical runtime analysis results has been presented in \cite{DBLP:conf/gecco/DoerrJWZ13}.
Fixed budget analysis can also be used to provide further insights of other types of randomized search heuristics.
For example, artificial immune systems have been analyzed using fixed budget analysis in \cite{DBLP:journals/tec/0001Z14}. The authors show how to translate previous runtime results on the use of hypermutations in artificial immune systems to results when given a fixed budget.

Related to fixed budget analysis is fixed target analysis~\cite{DBLP:conf/gecco/BuzdalovDDV20}. Instead of considering the solution quality obtainable within a given time budget, fixed target analysis considers the time needed to achieve a given target quality $k$. The runtime analysis is therefore parameterized by $k$. Results have been obtained for well studied benchmark problems such as OneMax and LeadingOnes. Furthermore, fixed target results for the classical minimum spanning tree problem have been obtained. In~\cite{AntipovBD20ppsn}, the dual question of how long an EA needs to find the optimum when already starting with a good solution was investigated. Surprisingly, different types of EAs profit to very different extents from such a favorable initialization. Also different parameter values become optimal when starting with a good instead of a random solution.

\subsection{Parameterized Analysis}
Parameterized analysis of algorithms~\cite{RodFell1999} allows to investigate algorithms not just with respect to the worst-case behaviour regarding the length of the given input, usually denoted by $n$, but also with respect to some additional parameter(s) that characterize the problem. 
A problem is defined to be fixed parameter tractable (FPT) with respect to a parameter $k$ iff there is an algorithm that runs in time $O(\poly(n) \cdot f(k))$, where $f(k)$ is a function only depending on $k$.
We call an algorithm an FPT algorithm with respect to a parameter $k$ iff it runs in time $O(\poly(n) \cdot f(k))$. This implies that an FPT algorithm runs in polynomial time if $k$ is constant.

The approach of analyzing evolutionary algorithms in the context of parameterized complexity has been introduced by Kratsch and Neumann~\cite{KratschN13} although there are earlier analyses that investigate RLS and the (1+1)~EA for the maximum clique problem on planar graphs~\cite{DBLP:conf/gecco/Storch06} and population-based EAs with respect to the runtime dependent on the size of the cliques obtained~\cite{DBLP:journals/tcs/Storch07}.
We call an evolutionary algorithm a fixed-parameter evolutionary algorithm with respect to a parameter $k$ iff its expected optimization time is $O(\poly(n) \cdot f(k))$. 
%As it is common in the runtime analysis of evolutionary algorithms, the expected optimization time refers to the expected number of fitness evaluations %until an optimal solution has been produced for the first time.

\subsubsection{Results for Minimum Vertex Cover}

\begin{algorithm2e}[t]
  \SetKwFor{For}{repeat}{}{}
  Choose an initial solution $x\in\{0,1\}^n$ uniformly at random\;
  Determine $f(x)$ and initialize $P \gets \{x\}$\;
  \For{forever}{%
    Choose $x\in P$ randomly\;
    Create $x'$ by flipping each bit of $x$ independently with probability $1/n$\;
    Determine $f(x')$\;
    \If{$\exists x'' \in P,\, f(x'')\leq f(x')$ and $f(x'')\not= f(x')$}{$P$ is unchanged}
    \Else{exclude all $x''$ where $f(x')\leq f(x'')$ from $P$
      and add $x'$ to~$P$}
  }
\caption{\gsemo for minimization problems}
\label{alg:paramcomplex:gsemo}
\end{algorithm2e}

The minimum vertex cover problem is the classical problem in the area of parameterized complexity and several FPT algorithms are available. 
The input of the minimum vertex cover problem is an undirected graph $G=(V,E)$ and the goal is to find a minimum set of vertices $V' \subseteq V$ such that each edge is covered by at least one node of $V'$, i.e. $e \cap V' \not = \emptyset$ holds for all $e \in E$.

Kratsch and Neumann~\cite{KratschN13} showed that a simple evolutionary multi-objective algorithm called \gsemo (see Algorithm~\ref{alg:paramcomplex:gsemo}) frequently used in the area of runtime analysis~\cite{DBLP:journals/nc/NeumannW06,DBLP:journals/ec/GielL10,FriedrichHHNW10} is able to compute a kernelization for the problem. A kernelization is a reduced problem where the decision for some nodes whether or not to include them has already been made in an optimal way. 
It is shown in \cite{KratschN13} that such a kernelization can be obtained by using two different types of helper objectives as a second objective. The first one considered in the article is the number of uncovered edges of a given solution $x$. The second approach uses the optimal value of the linear programming relaxation of the graph consisting only of the uncovered edges of a given solution $x$. Note that both helper objectives estimate the degree of infeasibility of a solution $x$ which is quite common when using multi-objective models for single-objective optimization problems in the context of evolutionary computing~\cite{DBLP:journals/nc/NeumannW06,FriedrichHHNW10}.

Having obtained such a solution an alternative mutation operator flipping bits corresponding to nodes that are adjacent to so far uncovered edges can obtain an optimal solution in time $O(f(k) \cdot \poly(n))$ which leads to the result that the examined evolutionary algorithms are fixed parameter evolutionary algorithms. It has also been shown that a factor $(1+\epsilon)$-approximation, $0 \leq \eps \leq 1$, can be obtained in expected time $O(n^2 \cdot \log n+ OPT \cdot n^2+n \cdot 4^{(1-\epsilon) \cdot OPT})$, where $OPT$ is the value of an optimal solution, when using the LP relaxation as the second objective. This gives a trade-off between approximation quality and runtime. Setting $\epsilon =1$, it shows that the approach computes a factor $2$-approximation in expected polynomial time.

In the weighted vertex cover problem, each node has a positive weight and the goal is to minimize the sum of the weights of the chosen nodes under the condition that all edges are covered.
The use of the dual formulation of the vertex cover in form of edge sets has been investigated by Pourhassan et al.~\cite{DBLP:journals/ec/PourhassanSN19}. They have generalized the edge based presentation by Jansen et al.~\cite{DBLP:conf/foga/JansenOZ13} to the weighted case and shown that their evolutionary multi-objective algorithm is a fixed parameter evolutionary algorithm for the weighted vertex cover problem.  The authors have shown that a $2$-approximation for the weighted vertex cover problem is obtained by the algorithm in expected polynomial time and
presented a population-based approach which achieves a $(1+\epsilon)$-approximation in expected time $O(n\cdot 2^{\min\{n, 2(1-\epsilon)OPT\}}+ n^3)$. Setting $\epsilon =1$, it shows that the approach computes a factor $2$-approximation in expected polynomial time for the weighted vertex cover problem.

\subsubsection{Results for the Euclidean Traveling Salesperson Problem}

The traveling salesperson problem is another very prominent problem in the area of combinatorial optimization. Given a set of $n$ cities $i=1, \ldots, n$, and distances $d(i,j)$ between them the goal is to compute a tour of minimal cost visiting each city exactly once and returning to the origin.
A possible solution for the TSP is usually given by a permutation $\pi=(\pi(1), \ldots, \pi(n))$ of the given $n$ cities and the goal is to find a tour $\pi$ that minimizes
$$
c(\pi) = d(\pi(n), \pi(1)) + \sum_{i=1}^{n-1}
d(\pi(i), \pi(i+1)).
$$

In the context of parameterized analysis of evolutionary algorithms, the Euclidean TSP has been investigated by Sutton et al.~\cite{DBLP:journals/ec/SuttonNN14}. Here each city $i$ is given as coordinates $(x_i, y_i)$ and the distance between city $i$ and $j$ is given as $d(i,j)=\sqrt{(x_j-x_i)^2 + (y_j-y_i)^2}$. The Euclidean TSP is still NP-hard but admits a polynomial time approximation scheme (PTAS)~\cite{DBLP:journals/jacm/Arora98}. In terms of parameterized analysis, the impact of the number of inner points has been considered which is given by the number of points that do not lie on the convex hull of the points in 2D. We denote by $n-k$ the points on the convex hull and $k$ the number of inner points.

The Euclidean TSP can be solved by classical algorithms in time $O(\poly(n) \cdot f(k))$ using dynamic programming~\cite{DBLP:journals/orl/DeinekoHOW06}. This makes use of the properties that an optimal solution has to visit the points of the convex hull in the order as they appear on the hull. The difficult task is then to "fill in" the inner points such that an optimal solution is obtained.

Investigations in the area of evolutionary algorithms focused on runtime analysis with respect to the number of inner points for the Euclidean TSP.
The first part of the analysis carried out in~\cite{Sutton2012tsp} analyzes the expected time until the classical (1+1)~EA using inversion as the mutation operators has computed a tour that is intersection free.
The analysis depends on the progress that can be made by inversion operations removing an intersection and this progress depends on the angle $\epsilon>0$ between any three points in the given set of cities. It is assumed that any three points are not collinear, i.e. do not lie on a single straight line.
For an intersection free tour it is known that the points on the convex hull appear in the permutation in the same order as they appear on the convex hull. The remaining part analyzes the expected time until an optimal solution has been produced from an intersection free tour. This part of the analysis uses that at most $2k$ inversion operations are sufficient to produce from an intersection free tour an optimal tour.
This implies that the (1+1)~EA obtains an optimal solution in expected time  $O(n^3 m^5 + n^{4k} (2k-1)!)$ when the points are placed on an $m \times m $ grid and no set of three points is collinear. Here the parameter $m$ for the grid directly determines the smallest angle that any set of three non collinear points can have. Note that the runtime bound does not meet the requirement of a fixed parameter evolutionary algorithm.

Afterwards, the ability of evolutionary algorithms to fill in the inner points correctly given that the points on the convex hull are in correct order has been examined. Ant colony optimization~\cite{DBLP:conf/cec/NallaperumaSN13a} and evolutionary algorithms~\cite{DBLP:conf/cec/NallaperumaSN13,DBLP:journals/ec/SuttonNN14} have been investigated in this context. For ant colony optimization, the crucial aspect to obtain a runtime of $O(n^k)$ is to construct solutions following the order on the convex hull. For evolutionary algorithms, a population-based algorithm building on a previous approach of Theile~\cite{Theile09} and  allowing to build an optimal tour following dynamic programming leads to a fixed parameter evolutionary algorithm with respect to the number of inner points. Furthermore, it is shown in \cite{DBLP:journals/ec/SuttonNN14} that a simple $(\mu+\lambda)$-EA searching for a permutation of the inner points and connecting them to the outer points using the dynamic programming approach given in \cite{DBLP:journals/orl/DeinekoHOW06} leads to a fixed parameter evolutionary algorithm.

\subsubsection{Further Results for Combinatorial Optimization Problems}
The early studies of Storch~\cite{DBLP:conf/gecco/Storch06} for the maximum clique problem in planar graphs investigated the runtime of RLS and the (1+1)~EA with respect to the size of the maximum clique. The fitness (to be maximized) of a search point $x \in \{0,1\}^n$, representing a selection of nodes, is given by the number of selected nodes if $x$ represents a clique and $-\infty$ otherwise. The algorithms investigated start with the initial solution $x=0^n$ which is a feasible solution.
For standard bit mutations an expected optimization time of $\Theta(n^6)$ has been shown for (1+1)~EA.
However, it should be noted that the size of a maximum clique in a planar graph is at most $4$ as the complete graph on $5$ vertices is not planar. Improved results have been shown in \cite{DBLP:conf/gecco/Storch06} for restart strategies used in RLS and for variants of the $(\mu+1)$-EA always deleting an individual with the worst fitness from the population.

The use of problem-specific mutation operators in the (1+1)~EA for the maximum leaf spanning tree problem has been investigated in \cite{DBLP:conf/ppsn/KratschLNO10}. In this work, it has been pointed out that standard bit flip mutations do not lead to fixed parameter evolutionary algorithms, where the parameter is the value of an optimal solution. Edge exchanges that include an edge currently not present in a spanning tree and that remove an edge from the resulting cycle are frequently used for spanning tree problems as they again lead to spanning trees.
Using edge exchanges for mutation where the number of edge exchanges is chosen according to a Poisson distribution with expected value $1$, it has been shown in \cite{DBLP:conf/ppsn/KratschLNO10} that the resulting (1+1)~EA is a fixed-parameter evolutionary algorithm when taking the value of an optimal solution $OPT$ as the parameter.

\subsection{Future work}
There is lot of potential for future studies using fine-grained analyses to provide new and important insights. Fixed budget and fixed target analysis has mainly been applied to classical benchmark problems but the number of results for classical combinatorial optimization problems is very limited. Obtaining fixed budget results for such combinatorial optimization problems which do not follow directly from the corresponding worst case analyses would be very valuable and show the quality obtainable within a given time limit. The same holds for fixed target analysis where such studies would reveal important insights for combinatorial optimization problems in terms of the time need to reach a desired quality.

The area of parameterized analysis of randomized search heuristics has been explored for some of the classical combinatorial optimization problems and such studies show which types of kernelization can be obtained in expected polynomial time. There are a wide range of kernelization techniques available in the classical parameterized complexity literature and it is unclear which types of kernels can be efficiently computed by evolutionary algorithms or similar search heuristics. Once a kernel is obtained, current studies use mutation operators that sample the reduced search space (almost) uniformly at random. While this leads to fixed parameter evolutionary algorithms, such operators are usually not that effective in practice. The question arises whether there are other useful techniques such as parameter adaptation that can be embedded into such mutation operators and ensure fixed parameter evolutionary algorithms and high performance in practice.

%\section{Fixed Budget and Fixed Target (??)}

\section{Setting the Parameters of Evolutionary Algorithms}
\label{sec:parameters}

The parameters of an evolutionary algorithm allow a user to adjust the EA to the problem to be solved and thus to optimize its performance. This is a great feature of EAs, but, at the same time, a difficult challenge~\cite{LoboLM07}. Missing good parameter values often results in horrible performance. Unfortunately, there is not much general advice on how to set the parameters. 

In this section, we show how theoretical works have helped to understand how the parameters of EAs influence their performance. Recently, the theory of EAs has also made big progress in understanding and even designing automated ways to find good parameter values. 

\subsection{Finding Optimal Static Parameter Values (Parameter Tuning)}

By \emph{parameter tuning} we understand the problem (or process) or finding suitable parameter values and then running the EA with these parameters. The parameter values are not changed during the run, so we speak also of \emph{static parameter values}. For reasons of space, we cannot discuss the whole literature on theoretical results that help tuning EA parameters, and therefore pick the mutation rate in elitist algorithms using standard bit mutation as the most prominent example. Other parameters that have attracted theoretical research include the parent and offspring population size (see, e.g, \cite{JansenJW05,Witt06,RoweS14,DoerrK15,AntipovDFH18}) and the selection pressure (see, e.g., \cite{JagerskupperS07,Lehre10,Lehre11,LehreY12,AntipovDY19}). For a discussion on how to set the parameters of estimation-of-distribution algorithms, we refer to Section~\ref{ssec:gdrift}.

We will focus our discussion in the following on the mutation rate, both because it is an important parameter and it is a parameter where a deep theoretical understanding exist. It is intuitively clear that we are faced with a trade-off situation when setting the mutation rate. A small mutation rate leads to slow a progress because the radius of exploration is small. A high mutation rate is detrimental because the random choice of the bits to be flipped on average increases the distance from the target solution, and this effect is linear in the mutation rate. 

An early established~\cite{Muhlenbein92,Back93} and generally accepted~\cite{Back96,BackFM97} recommendation is to use the mutation rate $p = \frac 1n$ in standard bit mutation, that is, we generate an offspring by flipping each bit independently with probability $\frac 1n$. With this choice, the expected distance between parent and offspring is one, so we inherit principles from local search. Different from local search, this mutation operator can leave local optima by flipping more than one bit. 

\begin{algorithm2e}[t]
  \SetKwFor{For}{repeat}{}{}
  Choose $x \in \{0,1\}^n$ uniformly at random\;
  \For{forever}{%
    \For{$i=1$ \KwTo $\lambda$}{
       Create $y_i$ by flipping each bit of $x$ independently with probability $1/n$\;
       }
    Choose $i \in [1..\lambda]$ such that $y_i$ has maximal fitness among $y_1, \dots, y_\lambda$, breaking ties uniformly at random\;
    \If {$f(y) \ge f(x)$}{$x := y$}
  }
\caption{The \oplea for maximization problems.}
\label{alg:oplea}
\end{algorithm2e}

A large number of mathematical runtime analyses shows that $p = \frac 1n$ often is optimal and thus complements the experimental support for this recommendation (see, e.g.,~\cite{Ochoa02} and the references therein). For the performance of the \oea on \onemax, a mix of rigorous and heuristic arguments already in~\cite{Muhlenbein92} and then fully rigorously in~\cite{GarnierKS99} shows that $p = \frac 1n$ is asymptotically optimal. For the \leadingones benchmark, a rate of $p \approx \frac{1.59}{n}$ was proven to be optimal in~\cite{BottcherDN10}. The \onemax result was greatly extended in~\cite{Witt13} with a proof that $p = \frac 1n$ is the asymptotically optimal mutation rate for each pseudo-Boolean linear function with non-zero coefficients. In~\cite{GiessenW17} it was proven that $p = \frac 1n$ is the asymptotically optimal mutation rate for the \oplea when the offspring population size $\lambda$ is not too large. The optimality of $p = \frac 1n$ was also shown for the optimization of long-path functions~\cite{Sudholt13}. For monotone functions, the situation is not fully understood, but again mutation rates around $p = \frac 1n$ appear to be a good choice. For the runtime of the \oea on strictly monotonically increasing functions, a $\Theta(n \log n)$ runtime can easily be shown when the mutation rate is $\frac cn$ for a constant $0 < c < 1$. That $\frac cn$ mutation rates for larger $c$ can lead to exponential runtimes was first shown in~\cite{DoerrJSWZ13}, the best known value for the constant~$c$ is $2.13...$~\cite{LenglerS18}. In the range around $p = \frac 1n$, for a long time only a runtime guarantee of $O(n^{3/2})$ was known for $p$ being exactly $\frac 1n$~\cite{Jansen07}. A significant progress on this long-standing problem was only made very recently -- in~\cite{LenglerMS19} an entropy compression argument was used to show that an $O(n \log^2 n)$ runtime guarantee holds for all mutation rates $p = \frac cn$, where $c \le c_0$ for some constant $c_0 > 1$. 

We note that the above results give some indication that $p = \frac 1n$ is a good first choice for the mutation rate, but by no means they prove that it always is. Indeed, already in~\cite{JansenW00} an example was constructed such that the \oea with any mutation rate that is not $\Theta(\frac{\log n}{n})$ needs super-polynomial time with high probability to optimize this problem. In~\cite{Prugel04}, the optimal mutation rates for the \oea optimizing hurdle functions with hurdle widths $2$ and $3$ were shown to be $\frac 2n$ and $\frac 3n$. This result could have led to the following findings, but apparently its broader implications on mutation rates (in a paper primarily discussing crossover) were not detected. So it was only in~\cite{DoerrLMN17} that the optimal mutation rate of the \oea on jump functions was shown to be roughly~$\frac kn$, where $k$ is the size of the fitness gap of the jump function. Also, it was shown that a small deviation from the optimal rate, say by a factor of $(1 \pm \eps)$, $\eps>0$ a constant, leads to a significant increase of the runtime by a factor exponential in~$k$. 

This result shows that the optimal mutation rate depends strongly on the input instance, that there is no rate that is universally good for all jump functions, and that the price for missing the right rate is significant. This lead the authors of~\cite{DoerrLMN17} suggest to use a random mutation rate, chosen independently for each mutation from a power-law distribution (we remark that random mutation rates were studied earlier to cope with unknown solution lengths~\cite{DoerrDK19} and higher-arity representations~\cite{DoerrDK18}). This heavy-tailed mutation operator shares with the classic mutation operator the property that a single bit (and more generally, any constant number of bits) is flipped with constant probability. When the power-law exponent is above two, then it also shares the property that an expected constant number of bits is flipped. Different from the classic recommendation, however, higher numbers of bits are flipped with larger probabilities. This essentially parameterless operator was shown to give on any jump function a performance of the \oea that differs from the one with instance-optimal mutation rate by only a small factor polynomial in~$k$. Heavy-tailed mutation operators proved to be successful in several other discrete optimization problems~\cite{FriedrichQW18,FriedrichGQW18,FriedrichGQW18heavysubm,WuQT18,AntipovBD20gecco,AntipovBD20ppsn,AntipovD20ppsn}. From a broader perspective, this line of work is an example showing that theoretical work not only can help understanding evolutionary algorithms, but it can also propose new operators and algorithms. 

\subsection{Dynamic Parameter Settings (Parameter Control)}\label{ssec:paracontrol}

Instead of trying to find a good parameter setting before starting the EA and sticking to this choice throughout the run of the EA, one could also think of optimizing the parameters during the run of the algorithm. This sophisticated-looking idea is called \emph{parameter control} and turns out to be less frightening than it appears at first. 

Indeed, the decision space (and thus also the opportunity to take an unsuitable decision) is much larger now -- in principle, we could choose different parameter values in each iteration -- but there are several powerful ways to overcome this difficulty. The advantage of parameter control is that we can react on the performance observed so far. This has two particularly positive consequences: (i)~The need for finding good parameter values before the start of the algorithm, based on a maybe only vague understanding of the problem to be solved, is reduced since a suboptimal initial choice can be corrected. (ii)~In the common situation that different parameter settings are optimal during different stages of the optimization process, we have the chance to use the optimal parameters for each stage (whereas a static choice would need to find a suitable trade-off). 

It is clear that the large space of different parameter settings for each iteration renders it unlikely to find the absolutely best dynamic choice of the parameters. However, it turns out that often very simple success-based or learning-based approaches lead to a very good performance, and often one that is better than the best static parameter setting. This is confirmed in many practical applications, see, e.g.,~\cite{KarafotiasHE15}, but also in now a decent number of theoretical works. 

The theoretical superiority of dynamic parameter settings over static ones was already demonstrated in~\cite{DrosteJW00} (see also~\cite{JansenW06} for an extension of this work), albeit for a simple algorithm with a simple \emph{time-dependent parameter choice} optimizing an artificial problem. Nevertheless, this result has rigorously proven that, in principle, dynamic parameter choices can efficiently solve problems where classic static choices would badly fail. Interestingly, the idea of time-dependent mutation rates was recently used again~\cite{RajabiW20} to help EAs leaving local optima.

It took ten years until dynamic parameter choices could be shown superior also for classic benchmark problems. The first such work~\cite{BottcherDN10} (see also~\cite[Section~2.3]{Doerr19tcs} for an extension) showed that a constant-factor runtime gain can be obtained from a \emph{fitness-dependent} choice of the mutation rate when optimizing the classic \leadingones benchmark via the \oea. Again it took some time until in~\cite{BadkobehLS14}, a super-constant runtime gain (of order $O(\log\log \lambda)$) from a dynamic parameter setting was shown for the \oplea optimizing \onemax. Other fitness-dependent parameter choices were discussed in~\cite{DoerrDE15,DoerrDY20}. A main problem with fitness-dependent parameter settings (or more generally speaking, parameter choices that depend on the current state of the algorithm) is that is needs a very good understanding of the problem to define a suitable functional dependence of the parameter value on the algorithm state. For the two examples from~\cite{BadkobehLS14,DoerrDE15}, it appears unlikely that without a mathematical analysis someone would have found the optimal functional dependence. Finding sub-optimal state-dependent parameter values that beat the best static values appears more realistic, but this remains a challenging task requiring a lot of expert knowledge.

Fortunately, there are dynamic parameter settings that need much less expertise. Generally speaking, these observe how the algorithm performs with the current parameter values (and sometimes also the values used in a longer history) and based on this try to adjust the parameter values to more profitable values. The easiest of these on-the-fly parameter choices are \emph{success-based multiplicative parameter updates}. Assume we suspect that increasing a specific parameter assignment also increases the chance to find an improvement, but at the price of increasing the computational cost of one iteration. Then increasing the current parameter value after each iteration without improvement and decreasing it after each iteration with improvement is a simple way to try to move the parameter value into a profitable region. Exactly this was suggested for the offspring population size $\lambda$ of the \oplea in~\cite{JansenJW05} and was rigorously analyzed in~\cite{LassigS11}, where an asymptotically optimal speed-up of the parallel runtime (number of iterations, ignoring the different costs of the iterations) was shown. The same basic idea was shown to give a (small) asymptotic improvement of the total runtime (number of fitness evaluations) for the \ollga optimizing \onemax~\cite{DoerrD18} and certain random SAT instances~\cite{BuzdalovD17}. 

The usual way to change the parameter value is multiplying or dividing by suitable constant factors. In~\cite{LassigS11}, the factor $2$ was used, and it is clear that any other constant factor would have given the same asymptotic runtime. In general, as observed in~\cite{DoerrD18}, smaller update factors can be the safer choice, and also the relation of the factors used in case of success and no success can be important. In~\cite{DoerrDL19},  a detailed analysis how the choice of these hyperparameters influences the runtime of the \oea with dynamic mutation rate on the \leadingones function was conducted. Other theoretical works on multiplicative parameter updates include~\cite{DoerrDK18} for multi-valued decision variables, \cite{MambriniS15} for migration intervals of island models, and \cite{DoerrLOW18} for the learning period of a hyperheuristic. We note that the results just described are the first examples of success-based parameter updates in discrete evolutionary optimization. In continuous optimization, a multiplicative update of the step size known as one-fifth rule was already proposed in~\cite{Rechenberg73}.

Multiplicative update rules work best if there is a simple monotonic influence of the parameter on the success, e.g., as seen for the offspring population size of the \oplea. Since such a simple relation is harder to find for the mutation rate in the \oplea, a different success-based scheme was developed in~\cite{DoerrGWY19}. Here half of the offspring are generated with twice the current rate, the other half with half the current rate. The mutation rate is then updated to the rate the best offspring was generated with (however, only with probability a half, with the other one-half probability the new rate is chosen randomly from the two alternatives). This mechanism was shown to let the \oplea optimize \onemax in asymptotically the same time as with the optimal fitness-dependent mutation rate developed in~\cite{BadkobehLS14}. 

A second way to go beyond multiplicative updates, and to additionally take more stable decisions, was proposed in~\cite{DoerrDY16ppsn}. Here for a small number of possible values of a parameter, a time-discounted estimate of the effectiveness of this parameter value was computed. In each iteration, with large probability, the best-performing value was used (exploitation) and with small probability a random one of the other values was used. With the right choice of the hyperparameters, this mechanism was shown to arbitrarily well approach the optimal mutation strengths of the \oea optimizing \onemax that were computed in~\cite{DoerrDY20}.

The most generic way to let an EA optimize its parameters itself is \emph{self-adaptation}, which means that the parameters are made part of the encoding of the solution candidates and thus become subject to variation and selection. The first to use this idea from evolution strategies in discrete evolutionary optimization was B\"ack~\cite{Back92ecal}. Taking the mutation rate as example, one appends an encoding of the mutation rate to the representation of the solution candidates. When mutating such an extended individual, one first mutates the mutation rate encoded in the extended individual and then, with the new rate, the remainder of the individual. The hope is that the suitability of a rate is visible from a higher fitness of the resulting individuals, and that the selection mechanisms of the EA bring these individuals (and thus the good mutation rate) forward in the population. While this way of adjusting parameters is clearly more natural for an EA than parameter adjustment mechanisms outside the evolutionary process, only three rigorous results supporting the usefulness of self-adaptation in discrete evolutionary computation have been published. In a first proof-of-concept work~\cite{DangL16ppsn}, an example is constructed that shows that self-adaptation can be useful. In this example, only two different mutation rates are available and it is assumed that the whole initial population starts in a particular search point. In~\cite{DoerrWY21}, the \oclea with self-adapting mutation rate is analyzed. With the hyperparameters suitably chosen, it can evolve sufficiently good mutation rates to obtain asymptotically the same performance on \onemax that was previously obtained with the optimal fitness-dependent setting~\cite{BadkobehLS14} and the two-population self-adjustment~\cite{DoerrGWY19}. A self-adaptive choice of the mutation rate for the \mclea optimizing \leadingones was proposed and analyzed in~\cite{CaseL20}. Here, the mutation rate is multiplied by a constant $A>1$ with some probability $p$, otherwise it is multiplied by a factor $b \in (0,1)$. Such a multiplicative update was used in a success-based manner in~\cite{DoerrDL19}. When the algorithm parameters $A, b, p, \mu, \lambda$ satisfy certain natural range constraints, then this self-adjusting algorithm optimizes the capped LeadingOnes function $\leadingones_k = \min\{\leadingones,k\}$ (defined on bit strings of length $n$) in time $O(k^2)$ for all $k \in [\Omega(\log^2 n)..n]$.

\subsection{The Future of Parameter Research}

The existing results show that we are now able to analyze a variety of static and dynamic parameter choices with a precision high enough to clearly distinguish good from bad choices. Some of these works not only analyzed existing algorithms or parameter adjusting mechanisms, but also suggested new approaches. Clearly, as true for all theoretical works, the algorithms and problems that were regarded are much simpler than those occurring in a practical application of EAs. To what extent the recommendations obtained from these simple settings generalize to more realistic ones is a crucial question which can only be answered in a collaboration between theoretical and applied researchers.

From the theory perspective, the following questions appear timely and interesting.
\begin{itemize} 
\item Interaction of parameters: So far, the vast majority of runtime analyses varies at most one parameter of the algorithm. Experience from practice shows that the interaction of several parameters is even harder to understand. So more runtime analyses discussing several parameters at once are clearly needed. Also, to the best of our knowledge, there is currently no theoretical work regarding two or more independent heavy-tailed parameters or self-adjusting or self-adaptive settings of two or more parameters. 
\item Self-adaptation: The most natural way to let an algorithm optimize its parameters is self-adaptation, where the parameters are integrated into the evolutionary cycle. So far, only very little theoretical advice exists how to successfully control parameters via self-adaptation. Here clearly more work is required.
\item Connections with machine learning: The area of machine learning has made tremendous progress in the last decades. Given that EAs are iterative algorithms in which often the state of the system changes only little in each iteration, one could envisage that dynamic parameter choices can profit from ideas and concepts borrowed from machine learning. While some ideas used in EAs can be related to similar ideas in machine learning, it seems to us that the full power of this connection has not yet been exploited. 
\end{itemize}

\section{Analysis of Evolutionary Algorithms in Dynamic and Stochastic Environments}
\label{sec:dynamic}

Dynamic and stochastic environments play a key role in real-world applications as information is often uncertain and circumstances change over time. Evolutionary algorithms have the ability to deal with changing circumstances and perform well in noisy environments which makes them well suited for dealing with dynamic and stochastic problems. A key contribution to the design of evolutionary for dynamic problems is the PhD Thesis of Branke~\cite{DBLP:books/daglib/0008593} which provides a wide range of techniques and investigations of evolutionary algorithms in dynamic environments.
Many dynamic problems have been tackled by evolutionary algorithms and other search heuristics. This includes important problems with frequent dynamic components such as vehicle routing~\cite{DBLP:journals/jco/MontemanniGRD05,DBLP:journals/apin/HansharO07}. 
Comprehensive presentations on the different techniques and problems investigated can be found in~\cite{DBLP:journals/swevo/NguyenYB12,DBLP:journals/swevo/MavrovouniotisL17}.

The area of runtime analysis has initially focused on simple toy problem in dynamic and stochastic settings. Again the function OneMax has played a crucial role to get initial insights. An important aspect in the context of dynamic optimization is how often and how drastic a function changes over time. We will describe important results for settings where the function or the constraints of a given problem change dynamically. Furthermore, we will summarize results where the fitness evaluation is impacted by noise and point out different results according to different noise models studied in the literature.
Additional investigations regarding dynamic and stochastic constraints in the context of submodular optimization are summarized in Section~\ref{sec:submodular}.

\subsection{Dynamic Benchmark Functions}

The runtime analysis for dynamically changing functions in discrete search spaces has been started by Droste~\cite{Droste02,DBLP:conf/gecco/Droste03}. He investigated a dynamic variant of the classical OneMax problem on binary strings. In the first dynamic setting, one randomly chosen bit is flipped in each iteration with probably $p$.
Droste~\cite{Droste02} showed that the expected optimization time of the (1+1)~EA is polynomial iff $p= O(\log(n)/n)$.
In the case where each bit is flipped in each iteration with a given probability $p$ investigated in~\cite{DBLP:conf/gecco/Droste03}, the runtime becomes super-polynomial if $p= \omega(\log(n)/n^2)$ and is polynomial if $p= O(\log(n)/n^2)$.
These investigations were revisited ten years later using drift analysis and generalized to the case where each element is not binary but can take on $r$ different values~\cite{DBLP:conf/foga/KotzingLW15}. 
Investigations on the magnitude and frequency of change for artificial benchmark functions have been carried out in~\cite{DBLP:conf/gecco/RohlfshagenLY09}. This study provides examples where the expected runtime changes drastically, i.e. from polynomial to exponential, if the frequency of change changes from low to high (or vice versa).

A comparison on the ability of simple evolutionary algorithms and ant colony optimization approaches for dealing with dynamic fitness functions has been carried out in~\cite{DBLP:conf/ppsn/KotzingM12,LissovoiW16}. These studies show that ant colony optimization can beat evolutionary algorithms due to their ability of adjusting slowly to changes in the fitness functions. Investigations of parallel evolutionary algorithms using island models carried out in \cite{DBLP:journals/algorithmica/LissovoiW18} for the MAZE function, introduced in~\cite{DBLP:conf/ppsn/KotzingM12}, show that infrequent migration of individuals is necessary for dense models where as infrequent migration becomes less necessary when working with sparse topologies in the island model. Furthermore, the usefulness of using non-elitist populations to track optima of dynamically changing problems has been pointed out in~\cite{DBLP:journals/algorithmica/DangJL17}.

\subsection{Dynamic Combinatorial Optimization Problems}

There are also some results on classical combinatorial optimization problems in the dynamic setting. Lissovoi and Witt~\cite{DBLP:journals/tcs/LissovoiW15} have investigated ACO algorithms and shown how the number of ants can impact different types of changes that can be tracked over time. They also give an example of dynamic oscillations that can not be tracked with a polynomial number of ants.
Dynamic makespan scheduling for two machines has been investigated by Neumann and Witt~\cite{DBLP:conf/ijcai/NeumannW15}. They have studied dynamic settings where solutions of small discrepancy of the two machines have to be recomputed. The results show that a worst case discrepancy of $U$, where $U$ is an upper bound on the maximal job length, can be maintained. Furthermore, better upper bounds on the runtime and lower discrepancies are shown for the case where the processing times of the jobs change randomly.

Dynamic variants of the minimum vertex cover problem have been considered in~\cite{DBLP:conf/gecco/PourhassanGN15,DBLP:journals/corr/abs-1903-02195}.
Following the edge-based encoding for the minimum vertex cover problem introduced in \cite{DBLP:conf/foga/JansenOZ13}, the problem formulation makes use of the dual formulation of the problem in order to represent solutions.
In \cite{DBLP:conf/gecco/PourhassanGN15}, the expected time to recompute $2$-approximation when edges are added or removed has been studied and improved results have been presented in~\cite{DBLP:journals/corr/abs-1903-02195}.

Dynamic settings of the classical graph coloring problem have been investigated in~\cite{DBLP:conf/gecco/Bossek0PS19}. Here, in particular, bipartite graphs have been studied and the necessity of complex mutation operators has been revealed even if there are only slight dynamic changes to the graph structure. These investigations have recently been extended in~\cite{DBLP:journals/corr/abs-2005-13825} and it has been shown that a dynamic setting where edges are presented to the algorithm in an iterative way can provably lead to better optimization times than presenting the algorithm with the whole input graph at once.

\subsection{Noisy Problems}
Studies in the area of runtime analysis of evolutionary computing techniques for discrete search spaces involving noisy objective functions have again been started by Droste~\cite{DBLP:conf/gecco/Droste04} who analyzed the (1+1)~EA on a noisy version of OneMax. He studied a prior noise model. In this case, some bits of a solution $x$ are flipped prior to the fitness evaluation. The studies considered flipping each bit with probability $p$ prior to fitness evaluation and Droste showed that the (1+1)~EA can still obtain the optimal solution for OneMax in expected polynomial time if $p=O(n/ \log n)$ whereas the expected optimization time becomes super-polynomial if $p=\omega(n/ \log n)$. In general, investigations can be separated into ones investigating prior noise as described above and posterior noise. In the case of posterior noise, the fitness of a solution $x$ is evaluated using the given fitness function $f$ but noise is added afterwards to the fitness value $f(x)$.
Gie{\ss}en and K{\"o}tzing~\cite{DBLP:journals/algorithmica/GiessenK16} build on this initial study by Droste and extended the study to population-based evolutionary algorithms and considered prior and posterior noise.
Results for prior bit-wise noise for the classical benchmark functions OneMax and LeadingOnes have been obtained in~\cite{BianQT18,QianBJT19}.
Additional and improved results including an example where noise helps have been provided by Sudholt~\cite{Sudholt18} and estimation of distribution algorithms (see Section~\ref{sec:eda}) have been studied for OneMax in \cite{FriedrichKKS17}. A method that can be used for the analysis of dynamic and noisy fitness functions has been developed in~\cite{Dang-NhuDDIN18}.
Furthermore, the trade-off between reducing noise through resampling and the computational cost of resampling has been investigated by Friedrich et al.~\cite{DBLP:conf/foga/0001KQS17}. The authors investigated classical evolutionary algorithms, estimation of distribution algorithms and ant colony optimization for OneMax and the case of additive posterior Gaussian noise. The use of non-elitist population based evolutionary algorithms for noisy versions of OneMax and LeadingOnes has been analyzed in~\cite{DBLP:conf/foga/DangL15}.

In terms of classical combinatorial optimization problems,
Sudholt and Tyssen~\cite{SudholtT12} investigated the stochastic single-destination shortest path problem and have pointed out settings where ant colony optimization is able to solve this problem in expected polynomial time or obtain a good approximation in expected polynomial time. These investigations have been extended in \cite{DoerrHK12ants} where a slight modification of the previously considered ant colony optimization approach has been analyzed and shown that this modified approach finds shortest path lengths efficiently although it does not necessarily converge.
Furthermore, stochastic constraints in form of chance constraints which require that stochastic constraints can only be violated with a small probability have been investigated recently (see Sections~\ref{sec:dynconst} and \ref{sec:subdyn}).

\subsection{Combinatorial Optimization Problems with Dynamic and Stochastic Constraints}
\label{sec:dynconst}

Dynamic constraints reflect the change in resources to solve a given problem. This is often a crucial aspect in many planning problems where resources such as trucks and trains might become unavailable due to failures or become available (again) after maintenance. Considering dynamic constraints, the objective function to be optimized is often assumed to be fixed and only changes to the constraints are considered.
The simplest example is the maximization of a linear function subject to a uniform constraint which limits the number of elements to be at most $B$. The first runtime analysis in this area considered the case where the bound $B$ changes to $B^*$ and the question is how long an evolutionary algorithm needs to recompute from an optimal solution for a given bound $B$ an optimal solution for the updated bound $B^*$. The (1+1)~EA and simple evolutionary multi-objective algorithms have been studied in~\cite{DBLP:journals/algorithmica/ShiSFKN19}. The multi-objective formulations enable to efficiently deal with constraint bound changes and this leads to significantly better experimental results on a wide range of knapsack instances as shown in \cite{DBLP:conf/ppsn/Roostapour0N18,DBLP:journals/corr/abs-2004-12574} if the magnitude and frequency of change is not too large. 

Chance constraints model constraints that are impacted by some noise of the given components and the goal is to optimize a given function $f$ under the condition that the constraint is violated with probability at most $\alpha$, where $\alpha$ is usually a small value, e.g.  $\alpha=0.001$.
Evolutionary algorithms for the chance-constrained knapsack problem have by studied from an experimental perspective by Yue et al.~\cite{DBLP:conf/gecco/XieHAN019,DBLP:journals/corr/abs-2004-03205,DBLP:conf/gecco/XieN021}. Furthermore, Assimi et al.~\cite{DBLP:journals/corr/abs-2002-06766} investigated evolutionary multi-objective evolutionary algorithms for the dynamic chance-constrained knapsack problem where the constraint bound for the knapsack dynamically changes over time through experimental studies.
A first runtime analysis for problems with chance constraints has been carried out by Neumann and Sutton~\cite{DBLP:conf/foga/0001S19} for special instances of the knapsack problem. It shows that even very simple linear functions with a simple linear stochastic constraint can lead to local optima with large inferior neighbourhoods that may make it hard for the (1+1)~EA to produce an optimal solution. These investigations have been extended by Yue et al.~\cite{DBLP:conf/gecco/XieN0S21} to the case of uniform stochastic weights where there are groups of elements that are correlated. For the setting in which every correlated group has the same profit profile, polynomial upper bounds on the expected optimization time of RLS and the (1+1)~EA have been given.

Important results on evolutionary algorithms for the optimization of submodular functions under dynamic and stochastic constraints have been obtained recently and are summarized in Section~\ref{sec:subdyn}. Furthermore, a more comprehensive and technical survey on the theory of evolutionary computing in dynamic and stochastic environments can be found in \cite{DBLP:journals/corr/abs-1806-08547}.

\subsection{Future work}
While there has been quite a large progress in recent years in this area of research, there are still a wide range of open problems and questions for future research.
Obviously, the understanding is still limited to basic benchmark functions and some well structured combinatorial problems and to further push the boundary of understanding is an important task.
Combining experimental analysis with mathematical investigations seems to be an important way to go as understanding evolutionary algorithms from a mathematical perspective is already challenging in static environments and even more true in the case of dynamic and stochastic problems. The design of evolutionary algorithms for dynamic and stochastic problems can (and has already been) guided by theoretical results. One example are dynamic and stochastic variants of the knapsack problem as investigated in \cite{DBLP:conf/ppsn/Roostapour0N18,DBLP:conf/gecco/XieHAN019} where the multi-objective formulations are based on previous theoretical results. Such interactions of experimental and mathematical research seems to be able to bridge the gap and allow to transfer theoretical results into the design of higher performing evolutionary algorithms.

\section{Submodular optimization}
\label{sec:submodular}

Submodular functions play a keyrole in the area of optimization. Many real world problems can be stated in terms of a submodular function as they face a diminishing return when adding additional components to a solution~(see the survey of Krause and Golovin~\cite{DBLP:books/cu/p/0001G14}). A wide range of greedy algorithms and local search approaches are available to solve different types of submodular problems with different types of constraints~\cite{DBLP:journals/mp/NemhauserWF78,DBLP:journals/ipl/KhullerMN99,DBLP:conf/stoc/LeeMNS09}. In the area of classical algorithms, these algorithms are generally designed and analyzed in terms of approximation guarantees that can be obtained in polynomial time~\cite{DBLP:journals/dam/ConfortiC84,vondrak2010submodularity}. The wider area of artificial intelligence also follows this approach but tests the performance of the developed approaches on important real-world problems matching the studied characteristics~\cite{DBLP:conf/aaai/ZhangV16,DBLP:conf/nips/Mitrovic0F0K19}. The previously mentioned studies can roughly be divided by the type of objective functions that are considered and the set of constraints. Furthermore, recent more general investigations consider the case where a function is not necessarily submodular and obtain results dependent on how close it is to being submodular measured in terms of the so-called "submodular ratio". Other important characteristics of the problems considered are noisy or dynamic problems~\cite{DBLP:conf/nips/Monemizadeh20,DBLP:conf/colt/HassidimS17} and problems where additional information is revealed over time~\cite{DBLP:journals/jair/GolovinK11,DBLP:journals/corr/abs-1911-03620}. 

Evolutionary algorithms have recently been analyzed and applied to different submodular optimization problems. We will summarize some of the key results in this section. For more details on some of the results we refer to the original articles or the recent book by Zhou et al.~\cite{DBLP:books/sp/ZhouYQ19}. This book gives a very comprehensive presentation on submodular optimization by evolutionary algorithms and discusses a wide range of submodular problems in the areas of optimization and machine learning.

We consider the following setting.
Given a set $V=\{v_1, \ldots, v_n\}$ of elements, the goal is to maximize a function $f\colon 2^V \rightarrow \mathds{R}^+$ that maximizes $f$ subject to a given set of constraints.
Submodular functions are usually considered in terms of marginal value when adding a new element.
We denote by $F_i(A)= f(A \cup \{i\}) - f(A)$
the marginal value of $i$ with respect to $A$.
A function $f$ is submodular
iff $F_i(A) \geq F_i(B)$ for all $A\subseteq B \subseteq X$
and $i\in X \setminus B$. Furthermore, a function $f$ is called monotone iff $f(A) \leq f(B)$ for $A \subseteq B$.

The first investigations in terms of runtime behaviour of evolutionary algorithms for submodular functions, we are aware of, have been carried out by Rudolph~\cite{DBLP:books/daglib/0017257} in the 1990s.
More than 15 years later this research area has been re-started by Friedrich and Neumann~\cite{2014PPSN_submod} and has since then gained significant attention.
The research in the context of static optimization can be grouped with respect to the type of objective functions and the type of constraints that are considered. In terms of objective functions, it is usually differentiated between monotone and non-monotone submodular functions. Furthermore, the submodularity ratio plays a crucial role when broadening the class of functions to functions that are not submodular. This ratio measures how close a function is to being submodular.
The submodularity ratio $\alpha_f$ of a given function $f$ is defined as

$$\alpha_f = \min_{X\subseteq Y,v\not \in Y}\frac{f(X\cup v) - f(X)}{f(Y\cup v)- f(Y)}\text{.}$$

Note that if $f$ is submodular then $\alpha_f=1$ holds.
The other important component in these investigations are the type of constraints that are considered. Constraints are usually of the type $c(X) \leq B$, where $c \colon 2^V \rightarrow \mathds{R}^+$ assigns a non negative cost to each set of elements $X \subseteq V$ and $B$ is a given constraint bound.
This type of constraints includes the case of a simple uniform constraints where $c(X) = |X|$ holds and limits the number of elements that can be included in a feasible solution by $B$. The maximization of a monotone submodular function under a uniform constraint is already NP-hard and can be approximated within a factor of $(1-1/e)$ by a simple greedy algorithm~\cite{DBLP:journals/mp/NemhauserWF78}.
More complex constraints involve partition or matroid constraints which are given in form of linear functions. Complex constraints that have been considered include cost values that can only be approximated, i.e. involving NP-hard routing problems. Then the approximation obtained for the submodular function depends on the type of objective function as well as the ability to calculate the cost of the considered constraint.
In the following, we summarize some of the main results in this currently very active research area.

\subsection{Monotone Submodular Functions}
\label{sec:monsub}

Optimal solutions for monotone submodular functions $f(X)$ with a cost constraint $c(X) \leq B$ can often be approximated well by simple greedy algorithms~(see~\cite{DBLP:books/cu/p/0001G14} for a comprehensive survey). 

Such greedy algorithms start with the empty set and add in each iterations an element with the largest marginal gain 
$$(f(X \cup\{x\}) -f(X))/(c(X \cup \{x\}) - c(X))$$
that does not violate the constraint.
The algorithm stops if no element can be added without violating the constraint bound.

Variants of \gsemo (see Algorithm~\ref{alg:paramcomplex:gsemo}) have been widely studied in the context of optimzing submodular functions. The initial analysis carried out in \cite{2014PPSN_submod} considered the maximization of monotone submodular functions with different types of constraints. After this \gsemo has been widely studied in the context of submodular optimization under the umbrella of Pareto Optimization which formulates a given constraint optimization problem as a multi-objective problem by establishing an additional objective based on the considered constraint. Such approaches have been widely used already before this in the context of runtime analysis of \gsemo. Solving single-objective problems by multi-objective formulations is a well-known concept in the evolutionary computation literature and has been studied from a practical and theoretical perspective since mid of the 2000s~\cite{DBLP:journals/jmma/Jensen04,DBLP:journals/nc/NeumannW06,DBLP:journals/tec/BrockhoffFHKNZ09}.

For the simplest case of a uniform constraint where $c(X) = |X|$, \gsemo can select in each step an element with the largest marginal again with respect to $f$. Friedrich and Neumann~\cite{2014PPSN_submod} have shown that \gsemo produces a $(1-1/e)$-approximation for monontone submodular functions with a uniform constraint in expected time $O(n^2(\log n +B))$ where $B\leq n$.
For monotone submodular functions with $k$ matroid constraints, local search and simple single-objective evolutionary algorithms such as the classical (1+1)~EA are able to obtain good approximation results. 
It has been shown by Lee at al.~\cite{DBLP:conf/stoc/LeeMNS09} that local search introducing at most $2p$ new elements and removing at most $2kp$ elements is able to obtain a $(1/(k + 1/p + \epsilon))$-approximation in polynomial time if $k\geq1$ and $p\geq 2$ are constants.
This result has later on been used to show that the (1+1)~EA is able to obtain the same approximation guarantee in expected time $O(\frac{1}{\epsilon} \cdot n^{2p(k+1)+1} \cdot k \cdot \log n)$. The crucial part of the proof is a result from ~\cite{DBLP:conf/stoc/LeeMNS09} which shows that every solution $x$ for which there is no $y$ in the defined neighborhood with $f(y) \geq (1 + \frac{\epsilon}{n(k+1)})\cdot f(x)$ is already a $(1/(k + 1/p + \epsilon))$-approximation.

Further investigations lead to a wide range of results for \gsemo on various submodular problems with cost constraints. The algorithm \gsemo is often called \pomc (Pareto Optimization algorithm for maximizing a monotone function with a monotone Cost constraint) (or similar) in such articles. The approaches are referred to as Pareto optimization and specify the class of problems to be tackled. However, usually the difference only lies in the formulation of the objective functions to formulate the constrained submodular problems as a multi-objective optimization problem. 

An important result covering a wide range of monotone functions for a broad class of cost constraints has been obtained by Qian et al.~\cite{DBLP:conf/ijcai/QianSYT17}. They investigated monotone functions in terms of submodularity ratio and general cost functions including ones for which it is hard to obtain an optimal solution exactly. 
Their theoretical results make use of proof ideas used for an adaptive greedy algorithm and show that \pomc is able to obtain the same approximation guarantee in expected pseudo-polynomial time. The expected runtime may be exponential with respect to the given input here if both the submodular function and the cost function can take on exponentially many values. In this case, the population size of \gsemo may become exponential during the run.
More precisely, they have shown that \pomc obtains a $(\alpha/2)\cdot (1 - e^ {-\alpha})$-approximation, $\alpha$ is the submodularity ratio, for a tightened cost constraint bound $\hat{B}$ (instead of B) where $\hat{B}$ depends on how well the given cost constraint can be approximated. Note that this setting includes problems where the cost of a solution may be hard to compute, i.e. for a selection of items it could be an approximation of a minimum Traveling Salesperson tour.
The experimental results show that \pomc clearly outperforms the adaptive greedy approach if the evolutionary algorithm is given a sufficient large number of fitness evaluations. Recently, an evolutionary multi-objective algorithm called EAMC has been introduced in \cite{bianefficient} which obtains the same worst-case approximation ratio as \pomc in expected polynomial time if the submodularity ratio of the given problem in known and used as a parameter in the algorithm. However, EAMC usually performs worse than \pomc on important benchmark problems.

Subset selection has also been investigated in the context of sparse regression. Here the submodular ratio $\alpha_f$ of the underlying function to be optimized plays a crucial role for the approximation quality obtained. Again, a variant of \gsemo called POSS~\cite{DBLP:conf/nips/QianYZ15} achieves in expected polynomial time the same approximation quality as a greedy approach called forward regression~\cite{DBLP:conf/icml/DasK11}, namely a solution $X$ with $f(x)\geq (1- e^{-\alpha}) \cdot OPT$. Furthermore, POSS outperforms forward regression and other simple heuristics in experimental investigations in terms of solution quality when giving it a sufficient amount of time to improve solutions during the evolutionary optimization process.

\subsection{Non-monotone Submodular Functions}

For symmetric functions which are not necessarily monotone and have $k$ matroid constraints, evolutionary algorithms and local search approaches can increase the function value by local operations to obtain a good approximation. Lee et al~\cite{DBLP:conf/stoc/LeeMNS09} have shown that if the value of a solution $x$ can not be increased by a factor of at least  $(1 + \epsilon/n^4)$ by changing at most $k+1$ elements, then $x$ is a  $\frac{1}{(k+2)(1+\epsilon)}$ -approximation. The series of such local improvements requires that the algorithm obtains a solution $x$ of value at least $f(x)\geq OPT/n$. Such a solution can be obtained from the empty set by adding the single element with the largest function value. Consequently local search algorithms building on such a solution and exchanging at most $k+1$ elements obtain a solution with the stated approximation quality in polynomial time~\cite{DBLP:conf/stoc/LeeMNS09}.
It has been shown that \gsemo obtains a $\frac{1}{(k+2)(1+\epsilon)}$ -approximation in expected time $O( (1/\epsilon) n^{k+6} \log n)$. The proof analyzes the process until a solution $x$ with $f(x)\geq OPT/n$ is obtained and the required number of local improvements until a solution of the stated approximation quality is obtained.

In their recent work, Qian et al.~\cite{DBLP:journals/ai/QianYTYZ19} give other major results which are broadening the setting of previous investigations. They considered an evolutionary multi-objective algorithm called \gsemoc which differs from \gsemo by producing from the offspring $x'$ a second offspring $x''$ which is the complement of the first offspring. The selection step of \gsemo is then applied to both $x'$ and $x''$. The authors first showed that for the case of non-monotone submodular functions without any constraint, \gsemoc is able to obtain a $(1/3 - \epsilon/n)$-approximation in expected time $O(\frac{n^4}{\epsilon}\log n)$. For $\epsilon$-monotone submodular functions, $\epsilon \geq 0$, where $f(X \cup \{x\}) \geq f(X)-\epsilon$ holds for any $X \subseteq V$ and $x \not \in X$, and a uniform constraint with bound $B$, they showed that \gsemoc achieves a solution $x$ with  $f(x) \geq (1 - 1/e) \cdot (OPT - k\epsilon)$  in expected time $O(n^2(B+ \log n))$ which generalizes the result given in \cite{2014PPSN_submod} to a wider range of functions by taking their closeness to monotonicity into account. 
Similar approximation results also hold for \gsemoc when considering $\epsilon$-approximately submodular functions, i.e. for functions $f$ for which a submodular function $g$ exists such that for all $X \subseteq V$, $(1-\epsilon)g(X) \leq f(X) \leq (1 + \epsilon)g(X)$ holds. The authors showed that suitable approximation can also be obtained for a wider range of functions with a cardinality constraint in expected time $O(n^2(B+ \log n))$. Specifically, they obtained results that depend on the submodularity ratio of the problem and investigated functions that are $\epsilon$-approximately submodular.

%\ignore{Greedy algorithms provide good approximation results when dealing with symmetric (non-)monotone %submodular problems involving matroid constraints~\cite{DBLP:conf/stoc/LeeMNS09}.}
Functions with bounded curvature under partition matroid constraints have been investigated in~\cite{DBLP:conf/aaai/000100QR19}. The results include the case of non-monotone submodular functions and approximation guarantees have been shown for the generalized greedy algorithm.
These investigations have recently been extended by Do and Neumann~\cite{VietPPSN20} to the evolutionary multi-objective algorithm \gsemo which is able to guarantee the same approximation quality as greedy but usually outperforms greedy in practice. 

\subsection{Submodular Functions with Dynamic and Stochastic Constraints}
\label{sec:subdyn}

Recent studies extended the investigations for monotone objective and costs functions to problems with dynamic constraints as well as constraints involving stochastic components.
Roostapour et al.~\cite{DBLP:conf/aaai/RoostapourN0019} investigated the setting of general cost constraints where the constraint bound $B$ changes over time. Generalizing the results of Qian et al.~\cite{DBLP:conf/ijcai/QianSYT17}  which are summarized in Section~\ref{sec:monsub}, they have shown that the evolutionary multi-objective approach \pomc computes an approximation for every budget $b$, $0 \leq b \leq B$. Furthermore, they have shown that if $B$ is increased to $B^*$, then an approximation for every $b$, $0 \leq b \leq B^*$ is obtained in pseudo-polynomial time. In contrast to this, it has been pointed out in \cite{DBLP:conf/aaai/RoostapourN0019} that simple adaptations of the generalized greedy algorithm are not able to maintain good approximations when dynamic changes are carried out. Furthermore, \pomc is able to learn the dynamic problems over time which gives it significant advantages over the greedy approaches as shown in comprehensive experimental investigations~\cite{DBLP:journals/corr/abs-1811-07806}.

Recently, the investigations in the area of submodular optimization have also been extended to stochastic constraints. Chance constraints play an important role in stochastic settings. These model situations where components of a constraint are stochastic and the goal is to optimize a given submodular objective function such that the probability of violating a given constraint bound is at most $\alpha$. Doerr et al.~\cite{DBLP:journals/corr/abs-1911-11451} investigated greedy algorithms for the optimization of monotone submodular functions for two settings. In the first setting, the stochastic weights are identically and independently uniformly distributed within a given interval $[a-\delta, a+\delta]$, $\delta \leq a$, where $\delta$ models the uncertainty  of the items. In the second setting each element $s$ has its own expected weight and is chosen independently of the others and uniformly at random in $[a(s)-\delta, a(s)+\delta]$, $\delta \leq \min_{s \in V} a(s)$. The investigations have recently been extended by Neumann and Neumann~\cite{AnetaPPSN20} to \gsemo and it has been shown that this algorithm is able to obtain the same approximation guarantee as the greedy approach in expected polynomial time in the case of identically and independently uniformly distributed weights. For the second setting, the same approximation guarantee as the one obtained for the greedy approach is obtained in expected pseudo-polynomial time. Furthermore, experimental investigations carried out for the influence maximization problem in social networks and the maximum coverage problem show that \gsemo significantly outperforms the greedy approach. A comparison of \gsemo to a standard setup of NSGA-II reveals that \gsemo is also often outperforming NSGA-II for the investigated settings which suggests that the ability of \gsemo to construct solutions in a greedy fashion is also crucial for the success of the algorithm in practice.

\subsection{Future work}
The studies show that evolutionary algorithms often provide the same worst case performance guarantees as classical algorithms but perform much better in practice. Many studies focus on variants of GSEMO which may encounter an exponential population size. Recent investigations show that special type of archived based algorithms such as EAMC only need a polynomial population size and achieve the same approximation guarantees. However, these algorithms usually perform worse in practice. Overall, it would be important to get a better understanding of classical evolutionary multi-objective algorithms for the use of Pareto optimization approaches in the context of submodular functions. Furthermore, the multi-objective model is usually used to enable a greedy process and the question arises whether the whole set of possible trade-offs in the multi-objective formulation is required to achieve good results from a mathematical and experimental perspective.
In terms of problem characteristics, it seems to be important to expand previous investigations, in particular, in the areas of dynamic, stochastic problems and further investigate the adaptation capabilities of evolutionary approaches for submodular problems in changing environments.

%The Benefits of Population Diversity in Evolutionary Algorithms: A Survey %of Rigorous Runtime Analyses

%\section{Other variants of evolutionary algorithms}

\section{Theory of Estimation-of-Distribution Algorithms}
\label{sec:eda}

Estimation-of-distribution algorithms (EDAs) are a more recent class of evolutionary algorithms (EAs). As a main difference to classic EAs, they do not evolve a population (that is, a finite set of solution candidates), but a probabilistic model of a solution candidate (that is, a probability distribution over the search space). Whereas a traditional EA selects individuals from a parent population, creates from them offspring via mutation and crossover, evaluates the offspring, and based on this evaluation selects from parents and offspring the next parent population, the EDA samples individuals from the current probabilistic model, evaluates them, and based on this evaluation defines the next probabilistic model. When viewing a parent population of a classic EA as probabilistic model (uniformly distributed on the individuals of the population), one can interpret population-based EAs as particular EDAs, but it is clear that the probabilistic models of EDAs are much more expressive than models building on finite populations. The obvious hope is that this richer class of algorithms contains better optimizers. However, there is also the additional hope that the probabilistic model evolved by an EDA can give insights beyond the good solutions that can be sampled from it. 

Most EDAs were defined in the 1990s, first in 1993 in an unpublished work~\cite{JuelsBS93} by Ari Juels, Shumeet Baluja, and Alistair Sinclair (see~\cite{Lobo07}) proposing the \emph{equilibrium genetic algorithm} (similar ideas can already be found in~\cite{Ackley87,Syswerda93}). While clearly containing the right ideas, this paper was never published and this algorithm is little known. Acknowledging the joint work with Ari Juels, Shumeet Baluja~\cite{Baluja94} proposed a very similar algorithm called \emph{population-based incremental learning (PBIL)}. As an important special case of it, M\"uhlenbein and Paass~\cite{MuhlenbeinP96} two years later suggested the \emph{univariate marginal distribution algorithm (UMDA)}. In 1999, Harik, Lobo, and Goldberg~\cite{HarikLG99} proposed the \emph{compact genetic algorithm (cGA)}. These and many other EDAs found numerous successful applications in the following years, see, e.g., the surveys~\cite{HauschildP11,LarranagaL02,PelikanHL15}.

First attempts to understand EDAs via theoretical means soon followed, starting -- as often -- with convergence results such as~\cite{HohfeldR97}. We note, however, that many of these very early results work with simplifying assumptions such as infinite population models and thus are not fully rigorous in the strict mathematical sense. In a series of works, Shapiro~\cite{Shapiro02,Shapiro05,Shapiro06} analyzed how the parameters of EDAs influence the effect of genetic drift. We discuss this central topic in more detail in Section~\ref{ssec:gdrift}. 

The first rigorous runtime analysis for an EDA was presented by Droste at GECCO 2005 (journal version~\cite{Droste06}). Chen, Lehre, Tang, and Yao~\cite{ChenLTY09} exhibited an artificial example problem which is easily solved by the UMDA, but for which the \oea with any $\Theta(\frac 1n)$ mutation rate needs exponential time to find the optimum. In~\cite{ChenTCY10}, Chen, Tang, Chen, and Yao discussed the use of frequency boundaries to prevent premature convergence. After these early works, it took another five years without theoretical works on EDAs until this area gained significant momentum in 2015--2016 with works like~\cite{DangLN19} (conference version at GECCO 2015), conducting a runtime analysis of the UMDA on \onemax and \leadingones, \cite{FriedrichKKS17} (conference version at ISAAC 2015) on the robustness of EDAs to noise, \cite{SudholtW19} (conference version at GECCO 2016) on how the update strength influences the runtime of the cGA, and~\cite{FriedrichKK16} pointing out that the main known EDAs are balanced, but not stable (that is, subject to genetic drift). These works generated a broad interest in theoretical analyses of EDAs, resulting is a large number of strong papers by a number of different authors. We refer to the recent survey~\cite{KrejcaW20bookchapter} for more details.

There is a natural connection between EDAs and ant colony optimization (ACO), since the pheromone system of an ACO algorithm leads to a probabilistic model of the search space as well. Two notable differences between classic EDAs and ACO algorithms are  (i)~that some ACO algorithms perform a pheromone update with the best-so-far solution, hence the state of the algorithm is not only described by the probabilistic model, and (ii)~that many ACO algorithms, in particular those for graph-based problems, use a very particular way of constructing new solutions. For this reasons, we do not discuss ACO algorithms in this section and refer the reader to the literature, e.g., the early concergence analyses~\cite{Gutjahr00,StutzleD02,Gutjahr02,Gutjahr03,SebastianiT05}, the first steps in runtime analysis for the 1-ANT algorithm and the MMAS with best-so-far improvement~\cite{NeumannW09,DoerrJ07cec,Gutjahr08,GutjahrS08,NeumannSW09,DoerrNSW11}, runtime analyses on graph-based problems~\cite{AttiratanasunthronF08,Zhou09,NeumannW10tcs,KotzingNRW10,KotzingLNO10}, runtime analyses for ACO algorithms with iteration-best improvement (which are very close to EDAs)~\cite{NeumannSW10,SudholtW19}, works demonstrating the robustness of ACO in stochastic environments\cite{KotzingM12,SudholtT12,DoerrHK12ants,FeldmannK13,LissovoiW16,FriedrichKKS16}, or the survey~\cite{Gutjahr11}, which still is a very valuable resource.

\subsection{The Compact Genetic Algorithm}

We now describe the compact genetic algorithm (cGA)~\cite{HarikLG99}, which will serve as a central example in this section. Other EDAs such as the UMDA or PBIL are substantially different, but appear to have similar strengths and challenges, so expecting similar results for these is a reasonable rule of thumb. However, we only concentrate on EDAs for discrete optimization problems here and we expect very different results in the continuous world. For reasons of simplicity, we only regard \emph{pseudo-Boolean} problems, that is, the optimization of functions $f : \{0,1\}^n \to \R$. 

The \emph{compact genetic algorithm} (cGA) is a \emph{univariate EDA}, that is, it treats the decision variables independently. For pseudo-Boolean problems, its probabilistic model is described by a \emph{frequency vector} $p \in [0,1]^n$. This frequency vector determines the following probability distribution on the search space $\{0,1\}^n$. If $X = (X_1, \dots, X_n) \in \{0,1\}^n$ is a search point sampled according to this distribution -- we write $X \sim \Sample(p)$ to indicate this -- then we have $\Pr[X_i = 1] = p_i$ independently for all $i \in [1..n] \coloneqq \{1, \dots, n\}$. In other words, the probability that $X$ equals some fixed search point $y$ is $\Pr[X = y] = \prod_{i : y_i = 1} p_i \prod_{i : y_i = 0} (1 - p_i)$.

In each iteration, the cGA samples two search points $x^1, x^2 \sim \Sample(p)$, computes their fitness, sorts them by fitness, that is defines $(y^1,y^2) \coloneqq (x^1,x^2)$ if $x^1$ is at least as fit as $x^2$ and $(y^1,y^2) \coloneqq (x^2,x^1)$ otherwise, and updates the frequency vector to $p \coloneqq p + \frac 1 K (y^1 - y^2)$, capped into the interval $[0,1]$, that is, with entries below zero replaced by zero and entries above one replaced by one. This definition ensures that when $y^1$ and $y^2$ differ in some bit position $i$, the $i$-th frequency moves by a step of $\frac 1 K$ into the direction of $y^1_i$ (but not below zero and above one). The \emph{hypothetical population size}~$K$, often also denoted by $\mu$, is an algorithm parameter that controls how strong this update is. To avoid a premature convergence, one often works with the \emph{frequency boundaries} $\frac 1n$ and $1-\frac 1n$, that is, one caps the new frequency vector into the interval $[\frac 1n, 1-\frac 1n]$ instead of $[0,1]$. 

This iterative frequency evolution is pursued until some termination criterion is met. Since we aim at analyzing the time (number of iterations) it takes to sample the optimal solution (this is what we call the \emph{runtime} of the cGA), we do not specify a termination criterion and pretend that the algorithm runs forever.
	
\begin{algorithm2e}%
	$p = (\frac 12, \dots, \frac 12) \in [0,1]^n$\;
	\Repeat{forever}{
    $x^1 \sim \Sample(p)$\;
    $x^2 \sim \Sample(p)$\;
    \leIf{$f(x^1) \ge f(x^2)$}{$(y^1,y^2) \assign (x^1,x^2)$}{$(y^1,y^2) \assign (x^2,x^1)$}
    $p \assign p + \frac 1 K (y^1-y^2)$ capped into $[0,1]$ or $[\frac 1n, 1-\frac 1n]$\;
  }
\caption{The compact genetic algorithm (cGA) to maximize a function $f : \{0,1\}^n \to \R$.}
\label{alg:cga}
\end{algorithm2e}

\subsection{Central Results}

In this section, we discuss three main insights which the theoretical analysis of EDAs has produced. For reasons of space, we point out two of them only briefly, namely that EDAs can perform well in noisy optimization and that they can cope well with local optima, and then discuss in detail how to set the parameters of EDA as this might the biggest obstacle in successfully using EDAs.

\subsubsection{EDAs Can Cope Well with Noise}

In their remarkable work~\cite{FriedrichKKS17}, Friedrich, K\"otzing, Krejca, and Sutton exhibit that the cGA is extremely robust to noise. More precisely, they show that the cGA with a suitable parameter choice can optimize a \onemax function subject to additive normally distributed noise in a runtime that only polynomially depends on the variance $\sigma^2$ of the noise. As they also show, such a performance cannot be obtained with many classic EAs. The reason for this robustness is the cautious update of the probabilistic model in each iteration (as opposed to the ``drastic'' alternatives of a classic EA, rejecting an offspring or keeping it and discarding some other individual). This caution of the EDA implies that a single wrong evaluation of a search point only has a small influence on the future run of the algorithm. In the only other study on how EDAs cope with noise, Lehre and Nguyen~\cite{LehreN19gecco} show that the UMDA with suitable parameter choices can optimize the \leadingones problem in time $O(n^2)$ also in the presence of constant-probability one-bit prior noise. 

\subsubsection{EDAs Can Cope Well with Local Optima}

Another difficulty for many EAs are local optima. Once the population of the EA is concentrated on the local optimum, it is difficult to leave this local optimum. As Hasen\"ohrl and Sutton~\cite{HasenohrlS18} (see also~\cite{Doerr19gecco}) show, the larger sampling variance of the cGA (in the regime without genetic drift) enables the algorithm to leave local optima much faster than many classic EAs. More specifically, Hasen\"ohrl and Sutton show that the cGA can optimize a jump function with jump size $k$ in time $\exp(O(k + \log n))$, whereas many mutation-based EAs need time $\Omega(n^k)$.

\subsubsection{Genetic Drift and Optimal Parameter Choices}\label{ssec:gdrift}

Both the result on noisy optimization and the one on local optima indicate that EDAs can have significant advantages over classic EAs. For reasons of brevity, we nevertheless omit further details and now turn to an important topic where a large sequence of works together have greatly increased our understanding, namely how to choose the parameters of EDAs and what is the role of genetic drift in EDAs.

While choosing optimal parameters for EAs is never easy, for many classic EAs a number of easy rules of thumb have been developed. For example, for mutation-based EAs the general recommendation to use standard-bit mutation with mutation rate $p = \frac 1n$ often gives reasonable results (though~\cite{DoerrLMN17} suggests that this impression is caused by an overfitting to unimodal problems). For EDAs, such general rules that are true over different classes of problems appear to be harder to find. From a large number of theoretical works, we now understand quite well why and we also have a number of different solutions to this problem. 

The main challenge is choosing an appropriate speed of adapting the probabilistic model. If this speed of adaptation is low, then it simply takes a long time to change the initial, usually uniform, model into a model that samples good solutions with reasonable probability. However, if the speed of adaptation is high, then the small random signals stemming from the random choices in the sampling of solutions are over-interpreted and the model is quickly adjusted to an incorrect model. When an EDA without frequency boundaries is used, this means that the model has (at least partially) converged to an incorrect model without the possibility to ever return. With frequency boundaries, there is still the chance to revert to a good model, but practical experience and theory shows (i)~that this can take a long time and (ii)~that usually the EDA continues to work with degenerate models and thus, to some extent, imitates classic EAs (and consequently does not profit from the more general model-building ability). The effect that frequencies without a justification from the fitness function move to boundary values is known as \emph{genetic drift}. 

Since genetic drift can lead to significant performance problems and since the risk of encountering genetic drift via unfortunate parameter choices is high, the question how to avoid genetic drift is, explicitly or implicitly, a common theme of almost all theoretical works on EDAs. Shapiro's very early works~\cite{Shapiro02,Shapiro05,Shapiro06} discussed this question explicitly, Droste's first rigorous runtime analysis regarded how the cGA optimizes \onemax only when the update strength $\frac 1K$ is $O(n^{-0.5 - \eps})$ for some constant $\eps > 0$, a parameter regime in which the cGA with high probability finds the optimum of \onemax in a way that never a frequency goes below $\frac 13$, that is, without encountering genetic drift. For reasons of space, we shall not describe in detail the whole history of understanding genetic drift of EDAs, but present immediately the final result only mentioning that both explicit investigations of genetic drift like~\cite{Shapiro02,Shapiro05,Shapiro06,FriedrichKK16,DoerrZ20tec} and the insights gained from many runtime analyses like~\cite{Droste06,DangLN19,LehreN17,SudholtW19,Witt19,KrejcaW20,LenglerSW21,HasenohrlS18,Doerr19gecco,Doerr19foga} paved the way towards this result.

Before discussing how to avoid genetic drift, let us quickly describe what is known about the danger of genetic drift. A first indication that genetic drift could be dangerous can be derived from the positive results -- the majority of the proven upper bounds for runtimes of EDAs only apply to regimes in which there is provably no genetic drift, and in fact, most proofs heavily exploit this. Rigorous proofs that genetic drift can lead to performance losses are much more rare and appeared only very recently, owing to the fact that lower bound proofs for EDAs are often very difficult. In their deep analysis~\cite{LenglerSW21}, Lengler, Sudholt, and Witt showed that the cGA with $K = \Theta(n^{0.5} / (\log n \cdot \log\log n))$ needs time $\Omega(n^{7/6}  / (\log n \cdot \log\log n))$ to optimize \onemax and the proof of this result shows that genetic drift is present. For $K = c n^{0.5} \ln n$, $c$ a sufficiently large constant, the cGA only needs time $O(n \log n)$ and here no genetic drift occurs~\cite{SudholtW19}. A more drastic loss from genetic drift, albeit on an artificial example problem, was observed in~\cite{LehreN19foga,DoerrK20evocop}. Lehre and Nguyen~\cite{LehreN19foga} define the deceiving-leading-blocks (DLB)\footnote{While not essential for the understanding of this section, for the sake of completeness we give a definition of the DLB problem. Let $n$ be even. Then the DLB problem with problem size $n$ is the following function $f$ defined on bit strings of length $n$. The unique global optimum of $f$ is $x^* = (1, \dots, 1)$, it has fitness $f(x^*) = n$. For all $x \in \{0,1\}^n \setminus \{x^*\}$, let $\LB(x) = \max\{i \in [0..n/2] \mid \forall j \in [1..2i] : x_j = 1\}$ denote the number of disjoint blocks of length two that are equal to $(1,1)$, counted from left to right until a different block is encountered. We call this block $(x_{2\LB(x)+1},x_{2\LB(x)+2})$ the critical block. The fitness $f(x)$ of $x$ is $2\LB(x)+1$, if the critical block equals $(0,0)$, and is $f(x) = 2\LB(x)$ if the critical block is $(1,0)$ or $(0,1)$.} problem and show that the UMDA with $\Omega(\log n) \le \mu = o(n)$ needs time exponential in $\mu$ to find the optimum. By~\cite{DoerrZ20tec}, in this parameter regime genetic drift is encountered when the runtime is $\omega(n^2)$. In~\cite{DoerrK20evocop}, it is shown that the UMDA with $\mu = \Theta(n \log n)$ can optimize the DLB problem in time $O(n^2 \log n)$ by profiting from the fact that now there is no genetic drift. A few experimental results also discuss the influence of genetic drift on the performance of an EDA, e.g., Figure~3 in~\cite{KrejcaW20bookchapter} shows the runtime of the UMDA on \onemax and Figure~1 in~\cite{DoerrZ20gecco} shows the runtimes of the cGA on \onemax, \leadingones, jump functions and the DLB problem. These results show a mild negative impact of genetic drift in the two \onemax experiments, a stronger impact for \leadingones, and a drastic impact for jump functions and DLB.

We now discuss how to predict and avoid genetic drift. A good way to measure genetic drift is by regarding a fitness function with a neutral bit, that is, a bit position that has no influence on the fitness. This might be overly pessimistic, since for such a bit the risk that the  frequency approaches an unwanted boundary value might be higher than for a bit with strong influence on the fitness, but (i)~a pessimistic view cannot be wrong here as a slightly too weak model update strength only slightly increases the runtime, whereas genetic drift as just seen can be detrimental, and (ii)~the results just described show that the estimates from regarding neutral bits, for these examples, cannot be far from the truth. 

The up to now most complete answer to the question of genetic drift was given in~\cite{DoerrZ20tec}, as said, a work that would not exist without the long sequence of previous works named above. We discuss this result in detail for the cGA and note that similar results are true for the UMDA and PBIL.

\begin{theorem}
  Let $f : \{0,1\}^n \to \R$. Assume that the $i$-th bit of $f$ is neutral, that is, $f(x) = f(y)$ for all $x, y \in \{0,1\}^n$ with $x_j = y_j$ for all $j \in [1..n] \setminus \{i\}$. Consider optimizing $f$ via the cGA with hypothetical population size $K$ using the frequency range $[\eps, 1-\eps]$ for some $\eps \in [0,\tfrac 14]$. Denote by $p^{(t)}$ the frequency vector resulting from the $t$-th iteration. 
  \begin{enumerate}
  \item Let $T^* = \min\{t \mid p_i^{(t)} \in \{\eps,1-\eps\}\}$. Then $E[T^*] = O(K^2)$.
  \item Let $T_{1/4} = \min\{t \mid p_i^{(t)} \in [0,\tfrac 14] \cup [\tfrac 34, 1]\} \le T^*$ be the first time the $i$-th frequency leaves the interval $(\tfrac 14, \tfrac 34)$ of the frequency range. Then $E[T_{1/4}] = \Omega(K^2)$. 
  \item\label{it:tail} For all $\gamma > 0$ and $T \in \N$, we have 
  $\,\Pr[\forall t \in [0..T] : |p_i^{(t)} - \tfrac 12| < \gamma] \ge 1 - 2\exp\left(-\frac{\gamma^2 K^2}{2T}\right)$.
%  \[\Pr[\forall t \in [0..T] : |p_i^{(t)} - \tfrac 12| < \gamma] \ge 1 - 2\exp\left(-\frac{\gamma^2 K^2}{2T}\right).\]
  \end{enumerate}
\end{theorem}

In very simple words the above result states that if we run the cGA for less than roughly $K^2$ iterations, then we do not encounter genetic drift, whereas after more than roughly $K^2$ iterations, genetic drift is likely to occur. 

The tail bound~(\ref{it:tail}) together with a simple union bound admits more precise guarantees, e.g., the following two formulations.
\begin{itemize}
\item If our aim is to run the cGA for $T$ iterations on some pseudo-Boolean function of dimension $n$, then by taking $K \ge \sqrt{32 T \ln(2n^2)}$ we can ensure that with probability at least $1 - \frac 1n$ no neutral bit has its frequency leave the interval $(\tfrac 14, \frac 34)$ within these $T$ iterations.
\item When $K$ is given, the probability that within $T \le \frac{K^2}{32 \ln(2n^2)}$ iterations a neutral frequency leaves the interval $(\frac 14, \frac 34)$ is at most $\frac 1n$. 
\end{itemize}
We note without further details that similar statements hold for bits which are not neutral, but which have a preference for a particular value $b$, that is, where changing the bit-value to $b$ can never decrease the fitness. Here the above statements hold for the undesired events that the frequency of this bit approaches the wrong boundary $1-b$. We refer to~\cite{DoerrZ20tec} for a precise statement of this result. This extension allows one to determine good values for the hypothetical population size for simple test functions like \onemax or \leadingones: If we run the cGA on one of these functions for $T$ iterations, then taking $K \ge  \sqrt{32 T \ln(2n^2)}$ ensures that with probability at least $1 - \frac 1n$ no frequency will go below $\frac 14$. 

For bit-values that have no uniform preference for a particular value (which is, naturally, the typical case for difficult optimization problems), we would still recommend to stick to the above-derived recommendations for setting $K$ since this at least avoids that frequencies reach the wrong value due to genetic drift. If a fitness landscape is strongly deceptive (the fitness drags the typical heuristics away from the global optimum), clearly, such arguments cannot avoid that frequencies approach the wrong end of the frequency range due to the misleading fitness signal. We note though that the heuristic argument for setting $K$ along the lines from above gives a good value and a good optimization behavior for the non-unimodal jump function class~\cite{HasenohrlS18,Doerr19gecco}. 

We finally note that there are three ``automated'' ways to approach the difficulty of finding the right parameter value. Inspired by the above insight, Doerr and Zheng~\cite{DoerrZ20gecco} proposed to start with a small value of $K$, run the cGA until either a satisfying solution is found or the time exceeds a limit up to which we are sure to not observe genetic drift and then restart with twice the $K$-value. In~\cite{Doerr19gecco}, a strategy is proposed that in parallel works with different $K$-values. Both approaches were proven to optimize simple test functions in a time that is by at most a logarithmic factor
larger than the runtime that can be obtained from using the optimal value of $K$. An experimental comparison~\cite{DoerrZ20gecco} gives no clear picture which of the two approaches is superior. Clearly, both perform better than what results from a static, but inappropriate choice of $K$. 

Instead of solving the genetic drift problem via a suitable choice of~$K$, the significance-based cGA proposed in~\cite{DoerrK20tec} tries to avoid genetic drift outright. We recall that genetic drift is caused by random fluctuations of the frequencies, which again are caused by sampling search points from the probabilistic model and updating the model based on these. Therefore,  the significance-based EDA avoids updating the model based on such short-sighted insights. Instead, this algorithm does not update the model until the history of the process gives sufficient evidence that some bit should better have a particular value. In this case, a drastic model update is performed by setting the corresponding frequency to $\frac 1n$ or $1 - \frac 1n$. This algorithm was shown to optimize both \onemax and \leadingones in time $O(n \log n)$, a performance not observed with any other classic EA or EDA so far. With no research on non-unimodal objective functions and no practical experience so far, of course, this is still a very preliminary line of research.

\subsection{Open Problems}

Being a very recent research topic, it is clear that the theory of EDAs contains more open problems than solved ones, and many open problems are fundamental for our understanding and the future use of EDAs. We first mention brief{}ly research topics where we feel that more results would greatly help and then give more details for two particular research questions.
%The following topics deserve more attention.
\begin{itemize}
\item Robust optimization: The only two results~\cite{FriedrichKKS17,LehreN19gecco} here show that the cGA can efficiently optimize \onemax in the presence of normally distributed additive posterior noise and that the UMDA can efficiently optimize \leadingones in the presence of one-bit prior noise. Having such results for other EDAs, other optimization problems, and other noise models (and other stochastic disturbances such as dynamically changing problem instances) would be highly desirable. 
\item Combinatorial optimization: While for classic EAs a large number of runtime analyses for combinatorial optimization problems exist~\cite{NeumannW10}, no such results have been shown for EDAs.% (the closest are results for ant colony optimizers).
\item Representations different from bit strings: For classic EAs, a number of results exist for problem representations different from bit strings, e.g.,~\cite{ScharnowTW04,DoerrJ10,DoerrDK18}, and these results show that the choice of the representation and the choice of the variation operators for these can make  a crucial difference. For EDAs, all results so far only discuss bit-string representations.
\end{itemize}
We now discuss in more detail two possible directions for future research.

\subsubsection{Runtime Results in the Regime with Genetic Drift} In the regime without genetic drift, EDAs often show a regular optimization behavior which often allows one to prove matching upper and lower bounds for runtimes. In the presence of genetic drift, the runtime is strongly influenced by how some frequencies approach the boundaries of the frequency range. It is thus rare events that determine the runtime and this makes it much harder to prove tight bounds. One could argue that runtime analyses in this regime are less interesting since we rather expect larger runtimes and rather an undesired behavior (e.g., imitating EAs), but this is not the full truth. For example, the UMDA optimizes \onemax in time $\Theta(n \log n)$ both for $\mu = \Theta(\log n)$ in the regime with (strong) genetic drift and for $\mu = \Theta(\sqrt n \, \log n)$ in the regime without genetic drift. 

Apart from sporadic results, which most likely are not tight in most of the cases, not much is known about the runtimes of EDAs in the genetic drift regime. In particular, the following questions are not understood.
\begin{itemize}
\item Runtimes of EDAs for very high update strengths: The few runtime analyses in the genetic drift regime all assume that the update strength is at least so small that a (sufficiently large) logarithmic number of frequency updates is necessary to bring a frequency to a boundary value, e.g., that $K \ge C \ln n$ for some sufficiently large constant $C$ when considering the cGA. Nothing nontrivial is known for even larger update strengths, but it is conjectured that one will typically encounter a super-polynomial runtime here.
\item Runtimes of EDAs on \onemax for moderate update strengths. For the case that the update strength is smaller than in the previous paragraph, but still high enough to lead to genetic drift, a general $\Omega(n \log n)$ lower bound for the cGA and UMDA~\cite{KrejcaW20,SudholtW19} and an $O(n \lambda)$ upper bound for the UMDA~\cite{DangLN19,Witt19} are known. For the cGA, a slightly stronger lower bound of $\Omega(K^{1/3} n)$ was shown in the regime $K = \Omega(\log^3 n) \cap O(\sqrt n / \log^2 n)$~\cite{LenglerSW21}. With this being all that is known, a true understanding of this regime is far from established.
\item Runtimes of EDAs on jump functions: In the regime without genetic drift, a reasonable understanding of the runtime of the cGA on jump functions has been obtained in~\cite{HasenohrlS18,Doerr19gecco,Doerr19foga}. The lower bound~\cite{Doerr19foga}, exponential in the jump size $k$, also applies to the regime with genetic drift, but is by far not sufficient to explain the huge runtimes observed experimentally~\cite{DoerrZ20gecco} in this regime. Hence a proof that the cGA optimizing jump functions suffers significantly from genetic drift is still missing.
\end{itemize}

\subsubsection{Multivariate EDAs}

While multivariate EDAs are used a lot and very successfully in practice, essentially all theoretical research so far regarded univariate EDAs only, that is, EDAs which evolve a univariate probabilistic model in which the bits are sampled independently. This is not surprising given how difficult it already was to obtain our limited understanding of univariate EDAs. 

So far, only two theoretical results on multivariate EDAs  exist, and both rely on theory-driven experiments and not on proven results. In~\cite{LehreN19foga}, the authors claim that the bivariate EDA \emph{mutual information maximization for input clustering (MIMIC)} can cope better with fitness landscape in which the decision variables are interdependent. They define an artificial fitness landscape with strong inter-variable dependencies, the DLB-problem, prove that the UMDA with $\mu = o(n)$ needs time exponential in $\mu$ and show experimentally that the MIMIC can optimize this landscape in time polynomial in $n$. Based on this finding, they suggest ``that one should consider EDAs with more complex probabilistic models when optimizing problems with some degree of epistasis and deception.'' As discussed in Section~\ref{ssec:gdrift}, the lower bound on the runtime of the UMDA only applies to the regime with strong genetic drift and from $\mu = \Omega(n \log n)$ on, the runtime of the UMDA on DLB becomes $O(\mu n)$~\cite{DoerrK20evocop}. For this reason, it is not clear if the MIMIC, and more generally, bivariate EDAs, are also superior to the UMDA with the right choice of the parameters. 

While so far no example exists in which a multivariate EDA was proven to have a better optimization behavior than a univariate one (with good parameters), the recent work~\cite{DoerrK20gecco} shows (again only experimentally) that bivariate EDAs can evolve very expressive probabilistic models. For a simple fitness landscape with $2^{n/2}$ global optima it is shown that the MIMIC very quickly evolves a probabilistic model which allows it to sample global optima with constant chance and in a way that very rarely an optimum is sampled repeatedly. Hence the model evolved indeed represents to some extend the structure of the set of optimal solutions. It is clear that this would not be possible with a univariate EDA or a population-based EA.

In summary, there is a cautious indication from theoretical work that multivariate EDAs could be interesting both from the viewpoint of good optimization times and good representations of the structure of the fitness landscape, but almost all of the work in this direction still needs to be done.

%Theoretical Foundations of Immune-Inspired Randomized Search Heuristics for %Optimization

%Computational Complexity Analysis of Genetic Programming

\section{Drift Analysis}
\label{sec:drift}

Drift analysis has become one of the most heavily employed tools in the mathematical analysis of evolutionary algorithms (EAs). Interestingly, it is one of the few tool sets which were not imported from the classic algorithms field. Rather, the classic algorithms field is now starting to use the drift theorems developed in our field, see, e.g.,~\cite{BertschingerLMMSTW20,GobelKK18,KosowskiU18,OgiermanE12}.

Using drift analysis as a tool in the performance analysis of EAs builds on the insight that it is often easy to estimate the expected progress (with regard to some suitable measure) of an EA in one iteration. Drift analysis therefore tries to translate this information into estimates for the first time that a particular goal is achieved. 

As a simple humorous example, inspired by a similar one from~\cite{Doerr11gecco}, consider the following question. You have an initial capital of \$1,000. Each day you go to your favorite pub and drink a random number of beers for an expected total price of \$10. After how many days you are bankrupt?

If there was no randomness involved, that is, if you would spend exactly \$10 each day, then obviously it takes exactly 100 days to spend your money. So does the answer change with randomness? Interestingly, it does not (of course, we can now only talk about the expected number of days to bankruptcy): The expected number of days until you have spend all your money is exactly 100, regardless of the distribution of the amount you spent per day (which could be different for each day, could depend on previous days, and could also take negative values). This is a simple application of the additive drift theorem (Theorem~\ref{thm:add} below). %If we start with $M$ dollars and spend an expected number of $d$ dollars a day (expected progress per iteration), then the expected number of days (iterations) until we reach the target state of bankruptcy is exactly $\frac Md$. 

The additive drift theorem is intuitive, but is in fact a deep mathematical result. Also, we have to note that it is not true that ``randomness never changes things''. Take for example the opposite process: You start with no money, but each day you earn an expected number of ten dollars. What is the expected time it takes until you have at least \$1,000? Now we can only say that it is at least 100 days (with a slightly less direct application of the additive drift theorem), but it could be much larger. For example, if each day we earn \$10,000 with probability $0.001$ and \$0 otherwise, then it takes an expected number of 1000 days until we have at least \$1,000.

Drift analysis was introduced to the field of evolutionary computation in the seminal paper~\cite{HeY01} of He and Yao, which builds on Hajek's fundamental work~\cite{Hajek82}. Since some of the early uses of drift arguments led to quite technical proofs, many researchers first shied away from using this new method and preferred classic arguments like Wegener's fitness level technique~\cite{Wegener01}. Over time, however, more elegant applications of the additive drift theorem, e.g., in~\cite{Jagerskupper08}, and drift theorems capturing better particular scenarios, e.g., the multiplicative drift theorem~\cite{DoerrJW12algo}, paved the way to drift analysis becoming perhaps the most powerful tool in the mathematical analysis of EAs.

\subsection{Three True Drift Theorems}

To show the beauty, simplicity, and power of drift analysis, we now present three central drift theorem. We call them \emph{true drift theorems} to reflect that all three translate information on the expected one-step progress into a hitting time without further assumptions on the distribution of the one-step progress. We state these theorems in their most basic version and trust that the reader is able to derive seemingly more general, but equivalent versions via scaling, shifting, or mirroring the random process.

\subsubsection{Additive Drift.}
From a deeper mathematical result of Hajek~\cite{Hajek82}, He and Yao~\cite{HeY01} derived the \emph{additive drift theorem} and used it to prove several runtime bounds. 

\begin{theorem}[additive drift theorem]\label{thm:add}
  Let $X_0, X_1, \dots$ be a sequence of random variables taking values in some finite set $S \subseteq \R_{\ge 0}$ with $0 \in S$. Let $T = \inf\{t \mid X_t = 0\}$. 
  \begin{itemize}
  \item Assume that there is a $\delta > 0$ such that for all $t \ge 0$ and $s \in S \setminus \{0\}$, we have $E[X_t - X_{t+1} \mid X_t = s] \ge \delta$. Then $E[T \mid X_0] \le \frac{X_0}{\delta}$.
  \item Assume that there is a $\delta > 0$ such that for all $t \ge 0$ and $s \in S \setminus \{0\}$, we have $E[X_t - X_{t+1} \mid X_t = s] \le \delta$. Then $E[T \mid X_0] \ge \frac{X_0}{\delta}$.
  \end{itemize}
\end{theorem}
Without going into details, we note that the assumptions can be weakened slightly, e.g., one can replace the ``point-wise drift requirement'', that is, the conditioning on $X_t = s$, by an ``average drift condition'', that is, conditioning only on $X_t > 0$~\cite{HeY01}. Also, the first part is also true for arbitrary infinite sets $S \subseteq \R_{\ge 0}$ and the second part is true also for bounded infinite sets $S$; see~\cite{LenglerS18}, where also a short and elegant proof of this result is presented.

The additive drift theorem gives good results if there is a roughly uniform progress regardless of time and state. In fact, as the two estimates together show, the additive drift theorem gives an exact estimate for the hitting time $T$ when the expected progress is known to be exactly $\delta$ at all times before hitting the target.

\subsubsection{Multiplicative Drift.}
For many natural optimization processes, the progress towards the optimum slows down when getting closer to the optimum. To use the additive drift theorem in such situations, the natural distance measure has to be transformed in such a way that the resulting expected progress is roughly uniform. Since the expected transformed progress is usually not just the transformation of the expected progress, such proofs can become technical and unintuitive.

Noting that a common situation is that the expected progress is roughly proportional to the distance to the target, in~\cite{DoerrJW12algo} a \emph{multiplicative drift theorem} was derived from the additive drift theorem. With a simpler direct proof, the following variant was later shown in~\cite{DoerrG13algo}. According to~\cite{Lengler20bookchapter}, the multiplicative drift theorem is the most often used drift theorem in the theory of evolutionary algorithms.

\begin{theorem}[multiplicative drift theorem]
  Let $X_0, X_1, \dots$ be a sequence of random variables over a state space $S \subseteq \{0\} \cup \R_{\ge 1}$ with $0 \in S$. Let $T = \min\{t \mid X_t = 0\}$. Assume that there is a $\delta > 0$ such that for all $t \ge 0$ and $s \in S \setminus \{0\}$, we have $E[X_{t+1} \mid X_t = s] \le (1-\delta) s$. Then the following estimates hold.
  \begin{itemize}
  \item $E[T \mid X_0] \le \frac{\ln(X_0)  + 1}{\delta}$.
  \item For all $\lambda > 0$, we have $\Pr[T > \lceil \frac{\ln(X_0) + \lambda}{\delta} \rceil] \le \exp(-\lambda)$.
  \end{itemize}
\end{theorem}

\subsubsection{Variable Drift.}
While indeed very many processes occurring in the analysis of evolutionary algorithms display an additive or multiplicative drift behavior, there remain processes in which the drift is decreasing when approaching the target (so that the additive drift theorem is hard to use), but not in a multiplicative fashion (so that the multiplicative drift theorem is hard to use). For these, so-called \emph{variable drift theorems} can be applied. The first variable drift theorem for the analysis of evolutionary algorithms was proposed by Mitavskiy, Rowe, and Cannings~\cite{MitavskiyRC09}, however, the independently developed result of Johannsen~\cite{Johannsen10} appears to be used more often. The following is a variant of Johannsen's result avoiding the use of integrals.

\begin{theorem}[variable drift theorem]
  Let $X_0, X_1, \dots$ be a sequence of random variables over a finite space $S$. Assume that $S = \{s_0, \dots, s_M\}$ with $0 = s_0 < s_1 < \dots < s_M$. Let $T = \min\{t \mid X_t = 0\}$. Assume that there is a monotonically non-decreasing function $h : S \setminus \{0\} \to \R$ such that for all $t \ge 0$ and $s \in S \setminus \{0\}$, we have $E[X_t - X_{t+1} \mid X_t = s] \ge h(s)$. Then $E[T \mid X_0] \le \sum_{i = 1}^{X_0} \frac{s_{i} - s_{i-1}}{h(s_i)}$.
\end{theorem}

The above are, most likely, the three most important drift theorems. We mention that the only other real drift theorem (that is, not requiring additional assumptions on the one-step distribution) we are aware of is the following result proven in~\cite{DoerrLO19}: Let a random process as in the multiplicative drift theorem be given, but with the drift condition $E[X_t - X_{t+1} \mid X_t = s] \ge \delta s$ replaced by the slightly stronger condition $E[X_t - X_{t+1} \mid X_t = s] \ge \delta s (\log_\gamma(s)+1)$ for some $\gamma > 1$. Then $E[T \mid X_0] \le \frac{3 + 4 \ln \gamma + \max\{0, 2 \log_2 \log_\gamma X_0\}}{\delta}$. We do not know if this result will find other applications, so we state the result here mainly to demonstrate that an only slightly stronger assumption on the drift -- $\Omega(s \log s)$ instead of $\Omega(s)$ -- can lead to a drastically smaller hitting time -- $O(\log\log X_0)$ instead of $O(\log X_0)$. 

\subsection{Drift Results With Additional Requirements}

The results presented in the previous section derive estimates for hitting times solely from the expected one-step progress; however, with two important restrictions: (i) except for the additive drift theorem, only upper bounds for hitting times can be obtained, and (ii)~only processes can be analyzed in which there is a drift towards the target that can be uniformly bounded or that decreases when approaching the target. 

Consequently, these drift theorems fail to describe a large number of behaviors of random processes that occur in the analysis of evolutionary algorithms. In this section, we brief{}ly describe such behaviors and what solutions for their analysis exist. Unfortunately, and this is the reason why we shall state no precise result, all these tools not only require information on the expected one-step change, but also on the distribution of the one-step change (typically, that the one-step change is concentrated around its expectation). For all results, this is not a weakness of the result, but an intrinsic necessity.

\subsubsection{Lower Bounds for Hitting Times.} From classic algorithms theory we know that it is very valuable to also have lower bounds on runtimes as these quantify how good our performance guarantees (upper bounds) are. If we have derived an upper bound from a certain drift behavior, say additive, multiplicative, or a certain variable drift, then the most natural approach would be to show a matching or near-matching upper bound on the expected one-step progress and derive (via a suitable drift theorem) from it a lower bound on the runtime. This works perfectly for the additive drift theorem as it contains such matching upper and lower bound results. 

For multiplicative and variable drift, the theorems presented in the previous section are missing such matching results, and this for good reason, namely because in general they are not true. As a simple example, consider the process on the state space $S = \{0,n\}$, starting with probability one in $X_0 = n$, which leaves state $n$ to $0$ with probability $1/n$ and stays in $n$ otherwise. Apparently, we have $E[X_{t+1} \mid X_t = n] = n-1 = (1 - \frac 1n) n$, that is, we have perfect multiplicative drift with $\delta = \frac 1n$. The multiplicative drift theorem thus gives an estimate for the expected hitting time of $E[T] = O(n \log n)$. This is best-possible in the sense that there are processes with multiplicative drift with $\delta = \frac 1n$ which indeed need $\Omega(n \log n)$ time, but for this particular process, the truth obviously is $E[T] = n$. This shows that a matching lower bound cannot exist without additional assumptions. 

The assumption that usually gives the desired behavior (and the desired lower bounds) is that the one-step progress is concentrated around its expectation, typically with some exponential tails or by forbidding large progresses at all. We spare the details and point the reader to~\cite{Witt13,DoerrDK18} for a multiplicative drift theorem for lower bounds and to~\cite{DoerrFW11,GiessenW18,DoerrDY20} for variable drift theorems for lower bounds.

\subsubsection{Increasing Drift.} All three main drift theorems require that the one-step progress is not increasing when approaching the target. This is a behavior often observed in evolutionary computation: The better the current solutions are, the harder it is to make progress. However, also the opposite behavior can be found, for example, when we consider how a better individual takes over a population. Here we would expect that the number of copies of the good individual increases in a multiplicative fashion (of course, only up to the point that a certain saturation is reached). Processes showing an increasing multiplicative drift have been analyzed in several papers dealing with population-based EAs, most notably in Lehre's~\cite{Lehre11} fitness-level approach to  non-elitist algorithms and in the so-called level-based theorem~\cite{CorusDEL18}. In both works, however, the increasing drift is mostly visible in the mathematical proofs. An explicit formulation of a drift result for increasing-drift processes was given in~\cite{DoerrK19}. Again, an expected multiplicative one-step progress is not enough, but some additional concentration assumptions are necessary. Motivated by the application to population processes, the additional assumption was made that the one-step progress stochastically dominates a binomial distribution. 

\subsubsection{Negative Drift.} A different situation is that a process shows a drift away from the target and that we want to argue that it takes a long time to reach this target. Such a situation naturally arises again in lower-bound proofs. The first such drift theorem was given by Oliveto and Witt~\cite{OlivetoW11,OlivetoW12}. Like many results proven later, see again the survey~\cite{Lengler20bookchapter}, it shows that if there is a constant negative expected progress in some interval of length $\ell$ and the one-step changes have both-sided exponential tails, then with probability $1 - \exp(-\Omega(\ell))$, the process takes time exponential in $\ell$ to reach the target. 

A different approach to analyze a negative drift situation was taken in~\cite{AntipovDY19}. Instead of the true process $X_t$, one regards an exponential transformation $Y_t = \exp(c (X_t - d))$ for suitable constants $c, d$, shows that $Y_t$ has at most a constant additive drift, and then uses the lower bound part of the additive drift theorem to derive the desired result. Depending on how easy it is to compute the drift of the transformed process, this approach might be technically simpler than using the existing negative-drift theorems. Different from all existing negative-drift theorems, it allows one to derive explicit constants in the exponent. As shown in~\cite{Doerr20gecco}, this approach can also give super-exponential lower bounds.

Very recently, a negative drift theorem without additional constraints was presented in~\cite{Doerr20ppsnLB}. At the moment, it is hard to foresee if it will find other applications than those presented in~\cite{Doerr20ppsnLB}.

\subsection{Challenge: Finding the Right Potential Function}

The results discussed so far show that we now have a decent number of drift theorems, which cover many different random processes. While surely new drift theorems will come up and existing ones will be polished, we are optimistic that the drift theorems developed in the last twenty years allow us to analyze most random processes occurring in the analysis of EAs. 

What is less understood, and often still a challenge, is defining the right random process. To be able to apply a drift theorem, we need to define a random process $(X_t)$ that describes some aspect of the run of our EA on some problem. Formally speaking, we need a function $g$ that maps the full state $S_t$ of the algorithm after iteration $t$ into a real number $X_t = g(S_t)$, and this in a way that the process $(X_t)$ still contains some relevant information of the run of the EA (e.g., that a suitable hitting time of $(X_t)$ corresponds to the time when an optimum was first found) and in a way that a drift theorem can be applied. While there are some generic solutions to this technical problem, many questions are still  open here and this might be the biggest challenge in the future of drift analysis.

A natural way to define the potential function $g$ is to take the fitness distance of the current-best solution to the optimum. This works well when there is a good correlation between the remaining optimization time and the fitness distance as observed, e.g., for the simple benchmarks \onemax and \leadingones (note that the classic analyses~\cite{DrosteJW02} stem from the time before drift analysis was introduced and hence use Wegener's~\cite{Wegener01} fitness level method) or combinatorial problems such as the minimum spanning tree problem (again, the classic proof~\cite{NeumannW07} does not use drift, but the expected multiplicative weight decrease method) or the maximum satisfiability problem with clauses of length~$3$~\cite{DoerrNS17}.

An equally natural potential is the structural distance to the optimum, e.g., the Hamming distance in the case of pseudo-Boolean optimization. This was used, e.g., to show that the \oea with mutation rate $c/n$, $c$ a constant strictly between $0$ and $1$, optimizes any strictly monotonic function in time $O(n \log n)$~\cite{DoerrJSWZ13}.

Unfortunately, these two potentials do not suffice. The most famous example for which neither the fitness distance nor the structural distance work well is the linear functions problem, that is, the question how fast the simple \oea optimizes a general linear pseudo-Boolean function $f(x) = \sum_{i=1}^n a_i x_i$. With a sequence of more powerful potential functions, all different from fitness and structural distance, increasingly strong results were obtained~\cite{DrosteJW02,DoerrJW12algo,DoerrG13algo,Witt13}.\footnote{We note that J\"agersk\"upper with a clever averaging argument could also use the structural distance as potential function.} Unfortunately, it remains unclear how to easily derive such potential functions. In fact, the only result regarding this question is a negative one, namely that to prove the results for larger mutation rates such as~\cite{DoerrG13algo,Witt13}, it is not possible to use one ``universal'' potential function for all linear functions, but the potential has to be chosen depending on the problem instance~\cite{DoerrJW12tcs}.

In three particular directions, we currently see the greatest lack of understanding how to define potential functions to use drift analysis. These are the following.

\subsubsection{Drift Analysis for Representations Other Than Bit Strings.} Once a relatively compact analysis of the runtime of the \oea with standard mutation rate $1/n$ on linear functions was found~\cite{DoerrJW12algo}, the question was raised how far these methods could be extended. One direction are linear functions defined not on bit strings, but on higher-arity representations $\{0, \dots, r\}^n$. While the $O(n \log n)$ runtime estimate could be shown for the search space $\{0,1,2\}^n$~\cite{DoerrJS11}, it was also shown in this work that there is no universal potential function from $r \ge 43$ on. With instance-specific potential functions, an $O(r n \log n + r^3 n \log\log n)$ upper bound was shown in~\cite{DoerrP12}. This extends the $O(rn \log n)$ bound to all $r = O((\frac{\log n}{\log\log n})^{1/2})$, but not beyond. It is an open problem whether larger $r$ indeed lead to an inferior runtime behavior or not. This example and the general shortage of works analyzing EAs with representations different from bit strings via drift analysis (we are only aware of~\cite{KotzingLW15,LissovoiW16,DoerrDK18}) suggest that more work is needed in this direction. 

\subsubsection{Drift Analysis for Population-based EAs.} All works described above, and in general the vast majority of runtime analyses building on drift arguments, only regard very simple EAs such as the \oea or, occasionally, the \oplea or the \opllga. For such EAs, a potential function only needs to estimate the quality of the single parent individual. For EAs working with a non-trivial parent population, it is much harder to define a suitable potential function. In fact, the main work on lower bounds for such algorithms by Lehre~\cite{Lehre10} used drift arguments only in the ancestral lines of single individuals and captured the effect of the whole population via family trees (see~\cite{Doerr20ppsnLB} for an alternative approach). Again for lower bound proofs, Neumann, Oliveto, and Witt~\cite{NeumannOW09} and later~\cite{OlivetoW15,AntipovDY19} used $\sum_{x \in P} c^{\onemax(x)}$ as potential (to be maximized) of a population $P$ in an algorithm maximizing the \onemax benchmark, where $c > 1$ is a suitable constant.  
The main line of works on upper bounds~\cite{Lehre11,DangL15,CorusDEL18} defines an extremely complicated potential function on the populations that is, most likely, not easy to transfer to problems not covered by these results. While these results show that drift analysis can be employed in the analysis of population-based algorithms, more work seems necessary, in particular, on upper bounds analyses, for drift analysis to replace the classic fitness level arguments more often employed here, see, e.g.,~\cite{Witt06,ChenHSCY09,AntipovDFH18}.

\subsubsection{Drift Analysis for Dynamic Parameters.} With the popularity of dynamic parameter choices both in theory (see also Section~\ref{ssec:paracontrol}) and practice, there is a strong need for mathematical methods to analyze such algorithms. From the perspective of drift analysis, again the challenge is to define a suitable potential function on the cross product of populations (which in the simplest case are just single individuals) and parameter values (or more generally, the full inner state of the algorithm). So far, we are only aware of the five works~\cite{DoerrDK18,AkimotoAG18,Rowe18,DoerrWY21,CaseL20} providing solutions to this problem. In the interest of brevity, we refer to~\cite[Section~1.3]{DoerrWY21} for a more detailed discussion, and state here only that our impression is that more work on this problem is necessary (and desirable) to ease future analyses of dynamic parameter settings.

\section{Final Words}
We provided an overview on areas of research in the field of theory of evolutionary computation in discrete search spaces that have gained significant attention during the last 10 years. The survey tried to capture the most important aspects from the perspective of the authors. We refer to the recent edited book~\cite{DoerrN20} for a more comprehensive overview, which also includes other evolutionary computing techniques such as genetic programming and artificial immune systems. For the true technical details, naturally, we invite the reader to consult the original articles. 

There are many areas where we see a lot of room for progress. Analyses for constrained problems static, dynamic, or stochastic have just recently been started and understanding the behavior of evolutionary algorithms for linear functions even very special simple constraints is still a challenging task~\cite{DBLP:conf/gecco/0001PW19}. A first theoretical analysis of differential evolution in discrete search spaces has been carried out in \cite{ZhengYD18}, however, indicating that our current methods cannot cope well with the complicated stochastic dependencies arising in this optimization process. The entropy compression method has found a first application in evolutionary computation~\cite{LenglerMS19}, but other applications of this powerful methods are not in sight. From a broader perspective, our understanding of the impact of populations, crossover operators, and diversity mechanisms still lags behind their practical success and proving the usefulness of such modules of an evolutionary algorithm for complex optimization problems is a challenging task.

We hope that the survey helps the readers to pursue their own research in this area. Although tremendous progress has been made during the last 10 years, there are still a lot of open questions and problems, some of which have been outlined in this article. We encourage the reader to make their own contribution to this field of research and help to transfer theoretical knowledge into the design of high performing evolutionary computing techniques.

\section*{Acknowledgements}
Frank Neumann has been support by the Alexander von Humboldt Foundation through a Humboldt Fellowship for Experienced Researchers and by the Australian Research Council through grant FT200100536.

\newcommand{\etalchar}[1]{$^{#1}$}

%\bibliographystyle{alphaurl}
%\bibliography{references,alles_ea,ich}

}%sloppy
\end{document}